\def\year{2022}\relax
\documentclass[letterpaper]{article} % DO NOT CHANGE THIS
\usepackage{aaai22}  % DO NOT CHANGE THIS
\usepackage{times}  % DO NOT CHANGE THIS
\usepackage{helvet}  % DO NOT CHANGE THIS
\usepackage{courier}  % DO NOT CHANGE THIS
\usepackage[hyphens]{url}  % DO NOT CHANGE THIS
\usepackage{graphicx} % DO NOT CHANGE THIS
\usepackage{natbib}  % DO NOT CHANGE THIS AND DO NOT ADD ANY OPTIONS TO IT
\usepackage{caption} % DO NOT CHANGE THIS AND DO NOT ADD ANY OPTIONS TO IT
\DeclareCaptionStyle{ruled}{labelfont=normalfont,labelsep=colon,strut=off} % DO NOT CHANGE THIS
\usepackage{algorithm}
\usepackage{algorithmic}
\usepackage{caption}
\usepackage{subfigure}
\usepackage{booktabs}
\usepackage{amsmath}
\usepackage{booktabs}
\usepackage{amsthm,amssymb}
\usepackage{newfloat}
\usepackage{listings}
\usepackage{multirow}
\usepackage{subfigure}
\usepackage{parskip}
\floatstyle{ruled}
\newfloat{listing}{tb}{lst}{}
\floatname{listing}{Listing}
\title{
ProtGNN: Towards Self-Explaining Graph Neural Networks
}

\title{ProtGNN: Towards Self-Explaining Graph Neural Networks}
\author {
    Zaixi Zhang$^1$,
    Qi Liu$^{1*}$,
    Hao Wang$^1$,
    Chengqiang Lu$^1$,
    Cheekong Lee$^2$\thanks{Qi Liu and Chee-Kong Lee are corresponding authors.}
}
\affiliations {
    \textsuperscript{\rm 1} Anhui Province Key Lab. of Big Data Analysis and Application, \\School of Computer Science and Technology, University of Science and Technology of China\\
    \textsuperscript{\rm 2} Tencent America\\
    \{zaixi, wanghao3, lunar\}@mail.ustc.edu.cn, qiliuql@ustc.edu.cn\\cheekonglee@tencent.com
}

\usepackage{bibentry}
\begin{document}

\maketitle

\begin{abstract}
Despite the recent progress in Graph Neural Networks (GNNs), it remains challenging to explain the predictions
made by GNNs. Existing explanation methods mainly focus on \emph{post-hoc} explanations where another explanatory model is employed to provide explanations for a trained GNN. The fact that \emph{post-hoc} methods fail to reveal the original reasoning process of GNNs raises the need of building GNNs with \emph{built-in} interpretability. In this work, we propose \textbf{Prot}otype \textbf{G}raph \textbf{N}eural \textbf{N}etwork (ProtGNN), which combines prototype learning with GNNs and provides a new perspective on the explanations of GNNs. In ProtGNN, the explanations are naturally derived from the case-based reasoning process and are actually used during classification. The prediction of ProtGNN is obtained by comparing the inputs to a few learned prototypes in the latent space.
Furthermore, for better interpretability and higher efficiency, a novel conditional subgraph sampling module is incorporated to indicate which part of the input graph is most similar to each prototype in ProtGNN+. Finally, we evaluate our method on a wide range of datasets and perform concrete case studies. Extensive results show that ProtGNN and ProtGNN+ can provide inherent interpretability while achieving accuracy on par with the non-interpretable counterparts. 
\end{abstract}
\section{Introduction}
Graph Neural Networks (GNNs) have become increasingly popular since many real-world relational data can be represented as graphs, such as social networks \cite{bian2020rumor}, molecules  \cite{gilmer2017neural} and financial data \cite{yang2020financial}. 
Following a message passing paradigm to learn node representations, GNNs have achieved state-of-the-art performance in node classification, graph classification, and link prediction \cite{kipf2016semi, velivckovic2017graph, xu2018powerful}.
Despite the remarkable effectiveness of GNNs, explaining predictions made by GNNs remains a challenging open problem. Without understanding the rationales behind the predictions, these black-box models cannot be fully trusted and widely applied in critical areas such as medical diagnosis. In addition, model explanations can facilitate model debugging and error analysis. These indicate the necessity of investigating the explainability of GNNs.

Recently, extensive efforts have been made to study explanation techniques for GNNs \cite{yuan2020explainability}. These methods can explain the predictions of node or graph classifications of trained GNNs with different strategies. For example, GNNExplainer \cite{ying2019gnnexplainer} and PGExplainer \cite{luo2020parameterized} are proposed to select a compact subgraph structure that maximizes the mutual information with the GNN's predictions as the explanation. PGM-Explainer \cite{NEURIPS2020_8fb134f2} firstly obtains a local dataset by random node feature
perturbation. Then it employs an interpretable Bayesian network to fit the local dataset and to explain the
predictions of the original GNN model. In addition, XGNN \cite{yuan2020xgnn} generates graph patterns to maximize the predicted probability for a certain class and provides model-level explanation. Despite the tremendous developments in the interpretation of GNNs, most existing approaches can be classified as \emph{post-hoc} explanations where another explanatory model is used to provide explanations for a trained GNN. \emph{Post-hoc} explanations can be inaccurate or incomplete in revealing the actual reasoning process of the original model \cite{rudin2018please}. Therefore, it is more desirable to build models with inherent interpretability where the explanations are generated by the model themselves.
%The fact that all these \emph{post-hoc} explanation methods do not provide the faithful reasoning process of GNNs raises the need of building GNN models with inherent interpretability.

We leverage the concept of prototype learning to construct GNNs with \emph{built-in} interpretability (i.e. self-explaining GNNs). In contrast to \emph{post-hoc} explanation methods, the explanations generated by self-explaining GNNs are actually used during classification and
are not generated \emph{post-hoc}. Prototype learning is a form of case-based reasoning \cite{kolodner1992introduction, schmidt2001cased}, which makes the predictions for new instances by comparing them with several learned exemplar cases (i.e. prototypes).  It is a natural practice in
solving problems with graph-structured data. For example, chemists identify potential drug candidates based on known functional groups (i.e. key subgraphs) in molecular graphs \cite{he2010predicting, zhang2021motif}. Prototype
learning provides better interpretability by imitating such a human problem-solving process. Recently the concept of the prototype has been incorporated in convolutional neural networks to build interpretable image classifiers
\cite{chen2018looks, rymarczyk2020protopshare}. However, so far prototype learning is not yet explored for
explaining GNNs.

Building self-explaining GNNs based on prototype learning brings unique challenges. First, the discreteness of the edges makes the projection and visualization of the graph prototypes difficult. Second, the combinatorial nature of graph structure makes it hard to build self-explaining models with both efficiency and high accuracy for graph modeling.

In this paper, we tackle the aforementioned challenges and propose \textbf{Prot}otype \textbf{G}raph \textbf{N}eural \textbf{N}etwork (ProtGNN), which provides a new perspective on the explanations of GNNs. Specifically, various popular GNN architectures can be employed as the graph encoder in ProtGNN. Prediction on a new input graph is performed based
on its similarity to the prototypes in the prototype layer. Furthermore, we propose to employ the Monte Carlo tree search algorithm \cite{silver2017mastering} to efficiently explore different subgraphs for prototype projection and visualization. In addition, in ProtGNN+, we design a conditional subgraph sampling module to identify which part of the input graph is most similar to each prototype for better interpretability and efficiency. Finally, extensive experiments on several real-world datasets show that ProtGNN/ProtGNN+ provides \emph{built-in} interpretability while achieving comparable performance with the non-interpretable counterparts.
\section{Related Work}
\subsection{Graph Neural Networks}
Graph neural networks have demonstrated their effectiveness on various graph tasks. Let $G = (V, E)$ denotes a graph with node attributes $X_v$ for $v \in V$ and a set of edges $E$. GNNs leverage the graph connectivity as well as node and edge features to learn a representation vector (i.e., embedding) $h_v$ for each node $v\in V$ or a vector $h_G$ for the entire graph $G$. Generally, GNNs follows a message passing paradigm, in which the representation of node $v$ is iteratively updated by aggregating the representations of $v$’s neighboring nodes $\mathcal{N}(v)$. Here
we use Graph Convolutional Network (GCN) \cite{kipf2016semi} as an example to illustrate such message passing procedures:
\begin{equation}
    h_v^{k+1} = \sigma \bigg(\mathop{\sum}_{u\in \mathcal{N}(v)}\big(W^{k}h_u^k\tilde A_{uv}\big)\bigg),
    \label{gcn}
\end{equation}
where $h_u^k$ is the representation vector of node $u$ at the $k$-th layer and $\tilde A = \hat D^{-\frac{1}{2}} \hat A \hat D^{-\frac{1}{2}}$ is the normalized adjacency matrix. $\hat A = A + I$ is the adjacency matrix of the graph $G$ with self connections added and $\hat D$ is a diagonal matrix with $\hat D_{ii} = \sum_j \hat A_{ij}$. $\sigma(\cdot)$ in Eq. (\ref{gcn}) is the ReLU function and $W^k$ is the trainable weight matrix of the $k$-th layer.

\subsection{Explainability in Graph Neural Networks}
As the application of GNNs grows, understanding why GNNs make such predictions becomes increasingly critical. Recently, the study of the explainability in GNNs is experiencing rapid developments. As Suggested by a recent survey \cite{yuan2020explainability}, existing methods for explaining GNNs can be categorized into several classes: gradients/features-based
methods \cite{baldassarre2019explainability, pope2019explainability}, perturbation-based methods \cite{ying2019gnnexplainer, luo2020parameterized, yuan2021explainability, schlichtkrull2020interpreting}, decomposition methods \cite{schwarzenberg2019layerwise, schnake2020xai}, and surrogate methods \cite{NEURIPS2020_8fb134f2, huang2020graphlime}.

Specifically, the gradients/features-based methods employ the gradients or the feature values to indicate the importance of different input features. These methods simply adapt existing explanation techniques in the image domain to the graph domain without incorporating the properties of graph data. Perturbation-based methods monitor the changes in the predictions by perturbing different input features and identifies the most influential features. Decomposition methods explain GNNs by decomposing the original model predictions into several terms and associating these terms with graph nodes or edges. %These methods generally follow a back-propagation manner to decompose predictions layer by layer until input space.
Given an input example, surrogate methods firstly sample a dataset from the neighborhood of the given example and then fit a simple and interpretable model, e.g., a decision tree to the sampled dataset.  The surrogate models are usually easier to interpret, shedding light into the inner-workings of more complex models.

However, all the above methods are \emph{post-hoc} explanation methods. Compared with \emph{post-hoc} explanation methods, \emph{built-in}
interpretability \cite{chen2018looks, ming2019interpretable} is more desirable since \emph{post-hoc} explanations usually do not fit the original model precisely \cite{rudin2018please}. Therefore, it is necessary to build models with inherent interpretability and high accuracy.
\section{The Proposed ProtGNN}
In this section, We introduce the architecture of ProtGNN/ProtGNN+, formulate the learning objective and describe the training procedures.
\subsection{ProtGNN Architecture}
\begin{figure*}[t]
	\centering
	\includegraphics[width=0.98\textwidth]{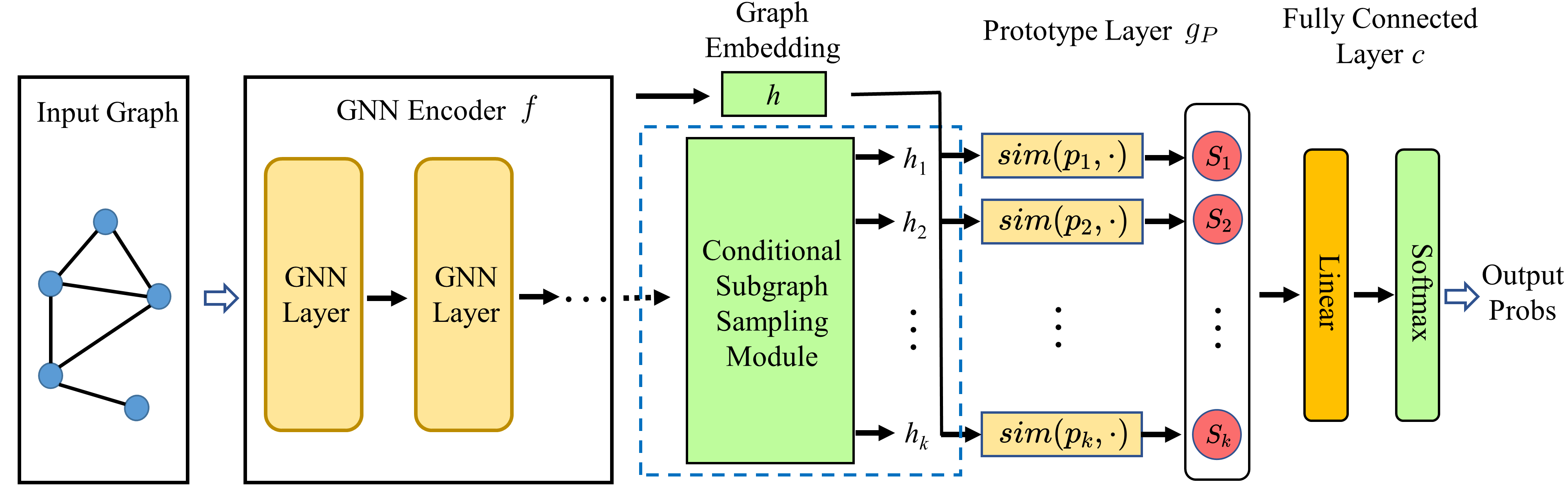}
	\caption{The architecture of our proposed ProtGNN/ProtGNN+. The model mainly consists of three parts: GNN encoder $f$, prototype layer $g_P$, and the fully connected layer $c$ appended by softmax to output probabilities in multi-class classification tasks. ProtGNN calculates the similarity score ($sim(p_k, \cdot)$ in the illustration) between the graph embedding and the learned prototypes in the prototype layer. For further interpretability, the conditional subgraph sampling module (in the dashed bounding box) is incorporated in ProtGNN+ to output subgraphs most similar to each learned prototype.}
	\label{protgnn+}
\end{figure*}
We let $\{x_i, y_i\}_{i=1}^n$ be a labeled training dataset, where $x_i$ is the input attributed graph and $y_i \in \{1,...,C\}$ is the label of the graph. We aim to learn representative prototypical graph patterns that can be used for classification references and analogical explanations. For a new input graph, its similarities with each prototype are measured in the latent
space. Then, the prediction of the new instance can be derived and
explained by its similar prototype graph patterns.

In Figure \ref{protgnn+}, we show the overview of the architecture of our proposed ProtGNN. The  network consists of three key components: a graph encoder $f$, a prototype layer $g_P$, and a fully connected layer $c$ appended by softmax to output the probabilities in multi-class classification tasks. 

For a given input graph $x_i$, the graph encoder $f$ maps the whole graph into a single graph embedding $h$ with a fixed length. The encoder could be any backbone GNN e.g., GCN, GAT or GIN. The graph embedding $h$ could be obtained by taking a sum or max pooling of the last GNN layer.

In the prototype layer, we allocate a pre-determined number of prototypes $m$ for each class. In the final trained ProtGNN, each class can be represented by a set of learned prototypes. The prototypes should capture the most relevant graph patterns for identifying graphs of each class. For each input graph $x_i$ and its embedding $h$, the prototype layer computes the similarity scores: 
\begin{equation}
    sim(p_k, h) = log(\frac{\|p_k-h\|_2^2 + 1}{\|p_k-h\|_2^2+\epsilon})
    \label{score function}
\end{equation} where $p_k$ is the $k$-th prototype with the same dimension as the graph embedding $h$. The similarity function is designed to be monotonically decreasing to $\|p_k-h\|_2$ and always greater than zero. In experiments, $\epsilon$ is set to a small value e.g., 1e-4. Finally, with the similarity scores, the fully connected layer with softmax computes the output probabilities for each class. 

\subsection{Learning Objective}
Our goal is to learn a ProtGNN with both accuracy and inherent interpretability. For accuracy, we minimize the cross-entropy loss on the training dataset: $\frac{1}{n}\sum_{i=1}^n{\rm CrsEnt}(c\circ g_p \circ f(x_i), y_i)$. For better interpretability, we consider several constraints in constructing prototypes for the explanation. Firstly, the cluster cost (Clst) encourages that each graph embedding should at least be close to one prototype of its own class. Secondly, the separation cost (Sep) encourages that each graph embedding should stay far away from prototypes not of its class. Finally, we found in experiments that some learned prototypes are very close to each other in the latent space. We encourage the diversity of the learned prototypes by adding the diversity loss (Div) which penalizes prototypes too close to each other.

To sum up, the objective function we aim to minimize is
\begin{equation}
    \frac{1}{n}\sum_{i=1}^n{\rm CrsEnt}(c\circ g_p \circ f(x_i), y_i) + \lambda_1 {\rm Clst} + \lambda_2 {\rm Sep} + \lambda_3 {\rm Div},
    \label{loss1}
\end{equation}
\begin{equation}
    {\rm Clst} = \frac{1}{n}\sum_{i=1}^n \mathop{\rm min}\limits_{j:p_j \in P_{y_i}} \|f(x_i) - p_j\|_2^2
\end{equation}
\begin{equation}
    {\rm Sep} = -\frac{1}{n}\sum_{i=1}^n \mathop{\rm min}\limits_{j:p_j \notin P_{y_i}} \|f(x_i) - p_j\|_2^2
\end{equation}
\begin{equation}
    {\rm Div} = \sum_{k=1}^C \sum_{\mathop{i \neq j}\limits_{p_i, p_j \in P_k}} {\rm max}(0, cos(p_i, p_j) - s_{max})
    \label{diversity loss}
\end{equation}
where $\lambda_1$, $\lambda_2$, and $\lambda_3$ are hyper-parameters controlling the weights of the losses. $P_{y_i}$ is the set of prototypes belonging to class $y_i$. $s_{max}$ is the threshold of the cosine similarity measured by $cos(\cdot, \cdot)$ in the diversity loss.

\subsection{Prototype Projection}
The learned prototypes are embedding vectors that are not directly interpretable. For better  interpretation and visualization, we design a projection procedure performed in the training stage.
Specifically, we project each prototype $p_j$ ($p_j \in P_k$) onto the nearest latent training subgraph from the same class as that of $p_j$ (see Eq. (\ref{projection equation})). In this way, we can conceptually equate each prototype with a subgraph, which is more intuitive and human-intelligible. To reduce the computational cost, the projection step is only performed every few training epochs:
\begin{equation}
\begin{aligned}
    p_j &\gets {\rm arg}~\mathop{\rm min}\limits_{\widetilde{h} \in \mathcal{H}_j}\|\widetilde{h}-p_j\|_2, \\
    \mathcal{H}_j=\{\widetilde{h}: f(\widetilde{x}), \widetilde{x}&\in {\rm Subgraph}(x_i)~\forall i~s.t.~y_i=k\}.
    \label{projection equation}
\end{aligned}
\end{equation}

\begin{figure}[t]
	\centering
	\includegraphics[width=0.48\textwidth]{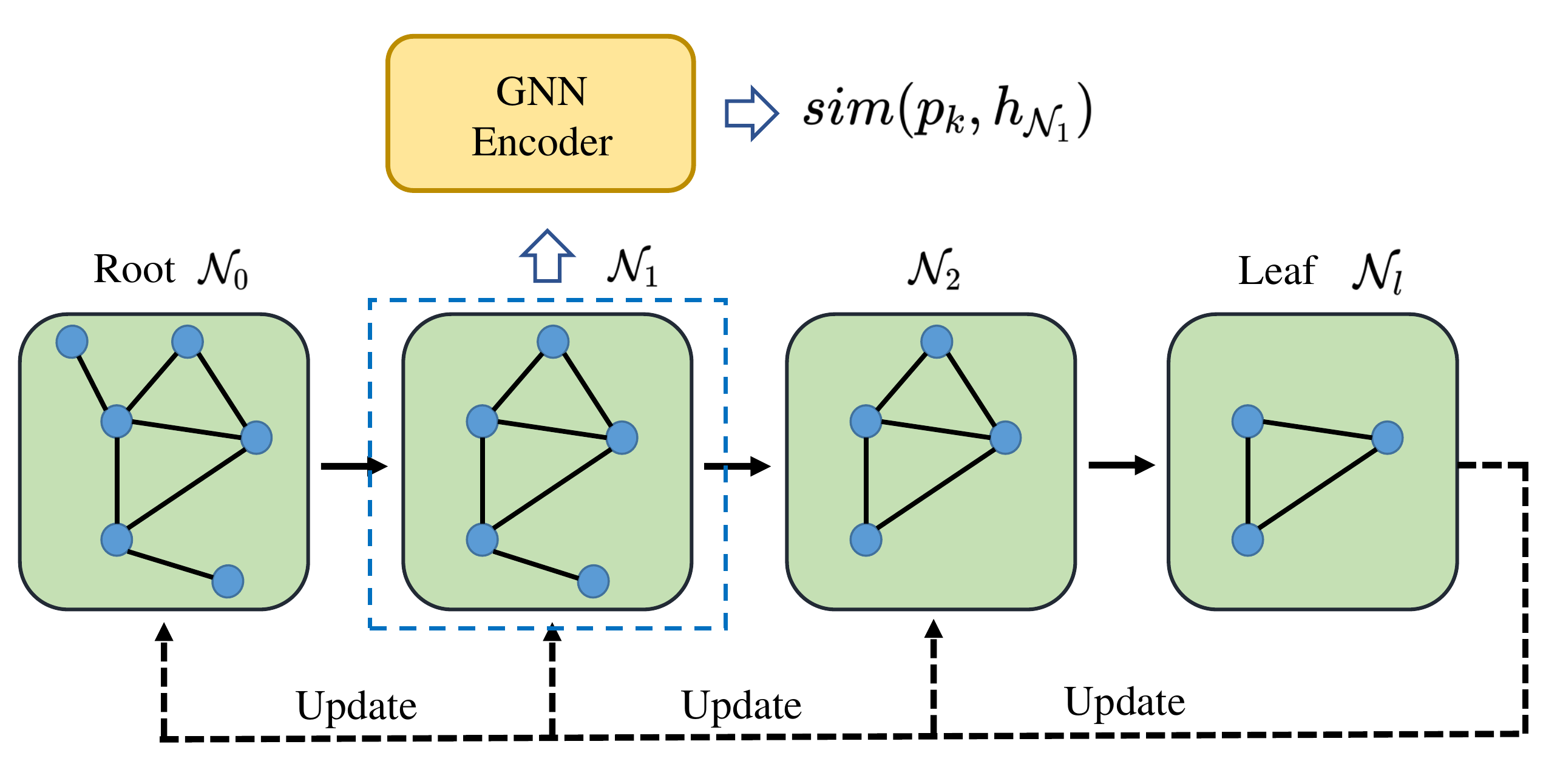}
	\caption{An illustration of graph prototype projection with Monte Carlo Tree Search. The bottom shows one selected path from the root to leaves in the search tree, which corresponds to one iteration of MCTS. Nodes that are not selected are ignored for simplicity. For each node, its subgraph is evaluated by computing the similarity score via GNN Encoder and the similarity function. In this figure, we show the computation of similarity score for the node $\mathcal{N}_1$ (shown in blue dashed box). In the backward pass, the model updates the statistics of each node.}
	\label{mcts}
\end{figure}

Unlike grid-like data such as images, the combinatorial characteristic of graph makes it unrealistic to find the nearest subgraph by enumeration \cite{chen2018looks}.
In graph prototype projection, we employ the Monte Carlo tree search algorithm (MCTS) \cite{silver2017mastering} as the search algorithm to guide our subgraph explorations (see Figure \ref{mcts}).  
We build a search tree in which the root is associated with the input graph and each of other nodes corresponds to an explored subgraph. Formally, we define each node in the search tree $\mathcal{T}$ as $\mathcal{N}_i$ and $\mathcal{N}_0$ denotes the root node. The edges in the search tree represent the pruning actions. In the search tree, the graph associated with a child node can be obtained by performing node-pruning from the graph corresponding to its parent node. To limit the search space,
we have added two additional constraints:  $\mathcal{N}_i$ has to be a connected subgraph and the size of the projected subgraph should be small.

During the search process, the MCTS algorithm records the statistics of visiting counts and rewards to guide the exploration and reduce the search space. Specifically, for the node and pruning action pair ($\mathcal{N}_i$, $a_j$), we assume that the subgraph $\mathcal{N}_j$ is obtained by action $a_j$ from $\mathcal{N}_i$. The MCTS algorithm records four variables for ($\mathcal{N}_i$, $a_j$):
\begin{itemize}
    \item C($\mathcal{N}_i$, $a_j$) denotes the number of counts for selecting action $a_j$ for node $\mathcal{N}_i$.
    \item W($\mathcal{N}_i$, $a_j$) is the total reward for all ($\mathcal{N}_i$, $a_j$) visits.
    \item Q($\mathcal{N}_i$, $a_j$) is the averaged reward for multiple visits.
    \item R($\mathcal{N}_i$, $a_j$) is the immediate reward for selecting $a_j$ on $\mathcal{N}_i$, which is measured by the similarity between the prototype and the subgraph embedding in this paper. The subgraph embedding is obtained by encoding the subgraph with the GNN encoder $f$. 
\end{itemize}
Guided by these statistics, MCTS searches for the nearest subgraphs in multiple iterations. Each iteration consists of two phases.
In the forward pass, MCTS selects a path starting from the root $\mathcal{N}_0$ to a leaf node $\mathcal{N}_l$. To keep subgraphs connected, we select to prune peripheral nodes with minimum degrees. The leaf node can be defined based on the numbers of nodes in subgraphs such that $|\mathcal{N}_l| \leq N_{\rm min}$. The action selection criteria at node $\mathcal{N}_i$ is:
\begin{equation}
    a^* = \mathop{\rm argmax}\limits_{a_j} Q(\mathcal{N}_i, a_j) + U(\mathcal{N}_i, a_j)
\end{equation}
\begin{equation}
    U(\mathcal{N}_i, a_j) = \lambda R(\mathcal{N}_i, a_j)\frac{\sqrt{\sum_k C(\mathcal{N}_i, a_k)}}{1+ C(\mathcal{N}_i, a_j)},
    \label{lambda}
\end{equation}
where $\lambda$ is a hyper-parameter to control the trade-off between exploration and exploitation. 
%This search strategy is a variant of the PUCT algorithm \cite{rosin2011multi}. 
The strategy initially prefers to select child nodes with low visit counts to explore different pruning actions, but asympotically prefers actions leading to higher similarity scores.

In the backward pass, the statistics of all node and action pairs selected in this path are updated: 
\begin{equation}
    C(\mathcal{N}_i, a_j) = C(\mathcal{N}_i, a_j) + 1
\end{equation}
\begin{equation}
    W(\mathcal{N}_i, a_j) = W(\mathcal{N}_i, a_j) + sim(p_k, h_{\mathcal{N}_l}),
\end{equation}
where $h_{\mathcal{N}_l}$ is the embedding of the subgraph associated to the leaf node $\mathcal{N}_l$.
In the end, we select the subgraph with the highest similarity score from all the expanded nodes as the new projected prototype.

\subsection{Conditional Subgraph Sampling module}
We further propose ProtGNN+ with a novel conditional subgraph sampling module to provide better interpretation. In ProtGNN+, we not only show the similarity scores to prototypes, but also identify which part of the input graph is most similar to each prototype as part of the reasoning process. In Figure \ref{protgnn+}, the conditional subgraph sampling module outputs different subgraph embeddings for each prototype. While this task can also be accomplished by MCTS, the exponentially-growing time complexity to the graph size and the difficulty of parallelization and generalization make MCTS algorithm an undesirable choice. Instead, we adopt a parameterized method for efficient similar subgraph selection conditioned on given prototypes.

Formally, we let $e_{ij} \in \{0, 1\}$ be the binary variable indicating whether the edge between node $i$ and $j$ is
selected. The matrix of $e_{ij}$ is denoted as $\mathcal{E}$. The optimization objective of the conditional subgraph sampling module is:
\begin{equation}
    \mathop{\rm max}\limits_{\mathcal{E}} sim(p_k, f(\mathcal{G}_s)) ~s.t. ~|\mathcal{G}_s| \le B,
    \label{optimization}
\end{equation}
where $\mathcal{G}_s$ is the selected subgraph whose adjacency matrix is $\mathcal{E}$. $B$ is the maximum size of the subgraph.

The combinatorial and discrete nature of graph makes the direct optimization of the above objective function intractable. We first consider a relaxation by assuming that the explanatory graph is a Gilbert
random graph \cite{gilbert1959random} where the state of each edge is independent to each other. Furthermore, for ease of gradient computation and update, we relax $\mathcal{E} \in \{0,1\}^{N \times N}$ into convex space $\mathcal{E} \in [0,1]^{N\times N}$. $N$ is the number of nodes in the input graph.
For efficiency and generalizability, we adopt deep neural networks to learn $e_{ij}$:
\begin{equation}
    e_{ij} = \sigma({\rm MLP}_{\theta} ([z_i;z_j;p_k])),
\end{equation}
where $\sigma(\cdot)$ here is the Sigmoid function. MLP is a multi-layer neural network parameterized with $\theta$ and [·; ·; ·] is the concatenation operation. $z_i$ and $z_j$ are node embedding obtained from the GNN Encoder, which encodes the feature and structure information of the nodes' neighborhood. Then the objective in Eq. (\ref{optimization}) becomes
\begin{equation}
\begin{aligned}
    \mathop{\rm max}\limits_{\theta} sim(p_k, f(\mathcal{G}_s)) - \lambda_b R_b&\\
    R_b= {\rm ReLU}(\sum_{e_{ij} \in \mathcal{E}} e_{ij} - B),
    \label{loss2}
\end{aligned}
\end{equation}
where $\lambda_b$ is the weight for the budget regularization $R_b$. In our experiments, we adopt stochastic gradient descent to optimize the objective function.

\textbf{Comparison with MCTS:} Our designed conditional subgraph sampling module is much more efficient than MCTS and easier for parallel computation. The parameters of our conditional subgraph sampling module are fixed and independent of the graph size. To sample from a graph with $|\mathcal{E}|$ edges, the time complexity of our method is $\mathcal{O(|E|)}$. One limitation of the conditional subgraph sampling module is that it requires additional training. Therefore, MCTS is still used in the prototype projection step of ProtGNN+ for the stability of optimization.
\subsection{Theorem on Subgraph Sampling}
To provide more understandable visualization, ProtGNN+ prunes the input graph to find the subgraphs most similar to prototypes and then calculates the similarity scores. Compared with ProtGNN, the subgraph sampling procedure may affect the classification accuracy. The following theorem provides some theoretical understanding of how input graph sampling affects classification accuracy.\\
\textbf{Theorem 1:} \label{theorem}Let $c\circ g_p \circ f$ be a ProtoGNN. The embedding of the input graph is $h$. We assume that the number of prototypes is the same for each class, and is denoted as $m$. For each class $k$, the weight connection in the last layer $c$ between a class $k$ prototype and the class $k$ logit is 1, and that between a non-class $k$ prototype and the class $k$ logit is 0. We denote $p_l^k$ as the $l$-th prototype for class $k$ and  $h_l^k$ the embedding of the pruned subgraph. ProtGNN and ProtGNN+ has the same graph encoder $f$.
We make the following assumptions: there exists some $\delta$ with $0<\delta<1$,
\begin{itemize}
    \item for the correct class, we have $\|h-h_l^k\|_2 \leq (\sqrt{1+\delta}-1)\|h-p_l^k\|_2$ and $\|h-p_l^k\|_2 \leq \sqrt{1-\delta}$;
    \item for the incorrect classes, $\|h-h_l^k\|_2 \leq \theta\|h-p_l^k\|_2 - \sqrt{\epsilon}$, $\theta = {\rm min}(\sqrt{1+\delta}-1, 1-\frac{1}{\sqrt{2-\delta}})$.
\end{itemize}
For one correctly classified input graph in ProtGNN, if the output logits between the top-2 classes are at least $2m log((1+\delta)(2-\delta))$, then ProtGNN+ can classify the input graph correctly as well.

The intuition behind Theorem 1 is that if the subgraph sampling does not change the graph embedding too much, ProtGNN+ will have the same correct predictions as ProtGNN. The proof is included in the appendix. 
\subsection{Training Procedures}
In Algorithm \ref{algorithm}, we show the training procedure of ProtGNN/ProtGNN+. Before training starts, we randomly initialize the model parameters. We let $w_c$ be the weight matrix of the fully connected layer $c$ and $w_c^{(k,j)}$ be the weight connection between the output of the $j$-th prototype and the logit of class $k$. In particular, for a class k, we set $w_c^{(k,j)} = 1$ for all $j$ with $p_j \in P_k$ and $w_c^{(k,j)} = 0$ for all $j$ with $p_j \notin P_k$. Intuitively, such initialization of $w_h$ encourages prototypes belonging to class $k$ to learn semantic concepts that are characteristic to class $k$. After training begins, we employ gradient descents to optimize the objective function in Eq. (\ref{loss1}). If the training epoch is larger than the projection epoch $T_p$, we perform the prototype projection step every few training epochs. Furthermore, if we train ProtGNN+, the conditional subgraph sampling module and ProtGNN are iteratively optimized after the warm-up epoch $T_w$ when the optimization of GNN encoder and prototypes are stabilized.

\begin{algorithm}[tb]
\caption{Overview of ProtGNN/ProtGNN+ Training}
\label{algorithm}
\textbf{Input}: Training dataset $\{x_i, y_i\}_{i=1}^n$\\
\textbf{Parameter}: Training epochs $T$, Warm-up epoch $T_{w}$, Projection epoch $T_p$, Prototype projection period $\tau$, ProtGNN+
\begin{algorithmic}[1] %[1] enables line numbers
\STATE Initialize model parameters. 
\FOR{training epochs $t = 1,2,\cdots ,T$}
\STATE Optimizing objective function in Eq. (\ref{loss1})
\IF {$t \textgreater T_p$ \textbf{and} $t \% \tau = 0$}
\STATE Performing prototype projection with MCTS
\ENDIF
\IF {ProtGNN+ enabled \textbf{and} $t \textgreater T_w$}
\STATE Optimizing the objective function in Eq. (\ref{loss2}).
\ENDIF
\ENDFOR
\\
\textbf{Output}: Trained model, prototype visualization
\end{algorithmic}
\end{algorithm}

\subsection{ProtGNN for Generic Graph Tasks}
In the above sections and illustrations, we have described ProtGNN/ProtGNN+ using graph classification as an example. It is worth noting that ProtGNN/ProtGNN+ can be easily generalized to other graph tasks, such as node classification and link prediction. For example, in the node classification task, the explanation
target is to provide the reasoning process behind the prediction of node $v_i$. Assuming the GNN encoder has $L$ layers, the prediction of node $v_i$ only relies on its $L$-hop computation graph. Therefore, prototype projection and conditional subgraph sampling are all performed in the $L$-hop computation graph.
\section{Experimental Evaluation}

\begin{figure*}[t]
	\centering
	\subfigure[]{\includegraphics[width=0.9\textwidth]{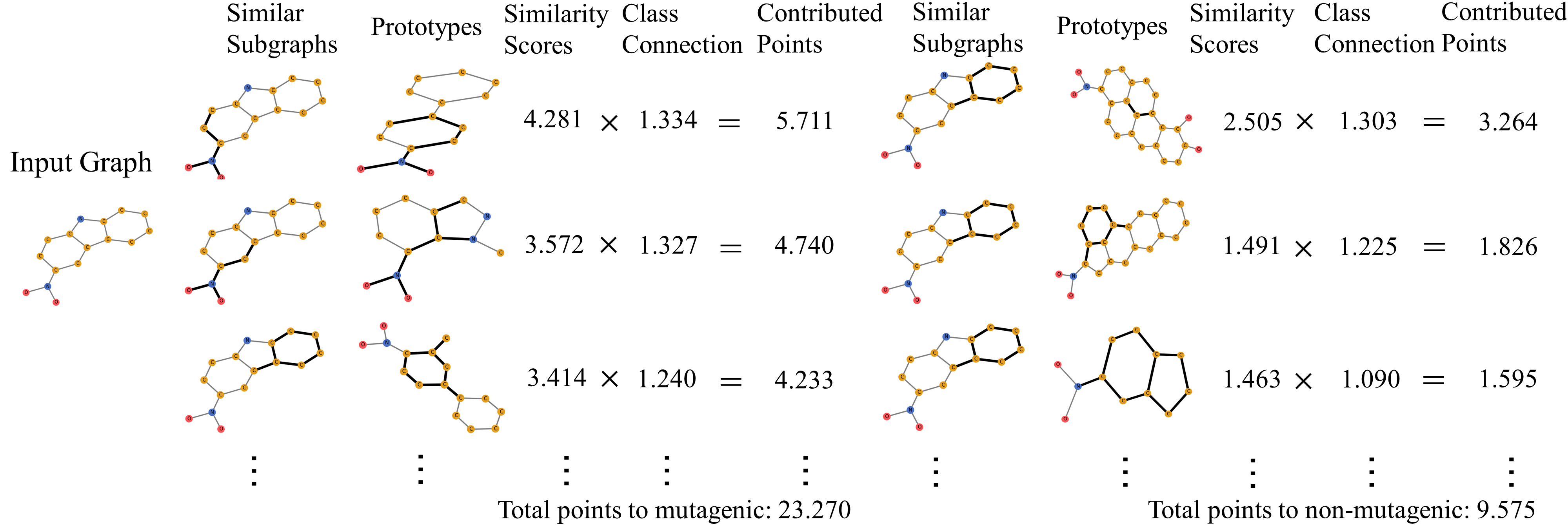}}
	\subfigure[]{\includegraphics[width=0.9\textwidth]{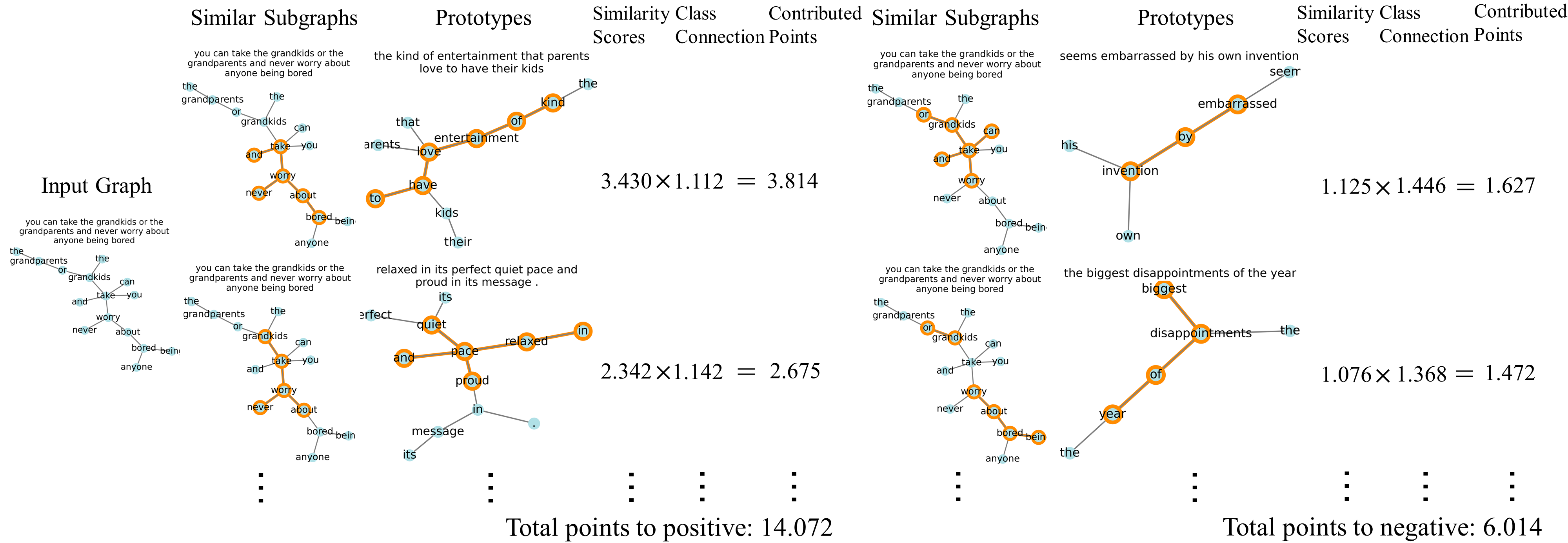}}
	
	\caption{The reasoning process of ProtGNN+ in deciding whether the molecular graph is mutagenic (a) or the sentiment of input text graph is positive (b).
	The predictions are based on the similarity between the latent input representations against the prototypes. The network tries to find evidence by looking at which subgraph was mostly similar to the prototypes. The selected subgraphs are highlighted. Due to space constraint, we only show several prototypes with the largest weights for each class.}
	\label{case study 1}
\end{figure*}
\subsection{Datasets and Experimental Settings}
\begin{table*}[t]
  \caption{The classification accuracies and standard deviations ($\%$) of ProtGNN, ProtGNN+, and the original GNNs. }
\small
\centering
\begin{tabular}{cccccccccc}
\toprule
\multirow{2}{*}{Datasets} & \multicolumn{3}{c}{GCN}       & \multicolumn{3}{c}{GIN}       & \multicolumn{3}{c}{GAT}       \\ \cmidrule(r){2-4} \cmidrule(r){5-7} \cmidrule(r){8-10}
                          & Original & ProtGNN & ProtGNN+ & Original & ProtGNN & ProtGNN+ & Original & ProtGNN & ProtGNN+ \\ \midrule
MUTAG                     & 73.3$\pm 5.8$       & \textbf{76.7}$\mathbf{\pm 6.4}$      & 73.3$\pm 2.9$        & \textbf{93.3}$\mathbf{\pm 2.9}$        & 90.7$\pm 3.2$       & 91.7$\pm 2.9$        & 75.0$\pm 5.0$        & 78.3$\pm 4.2$       & \textbf{81.7}$\mathbf{\pm 2.9}$        \\
BBBP                      & 84.6$\pm 3.4$        & \textbf{89.4}$\mathbf{\pm 4.1}$       & 88.0$\pm 4.6$        & 86.2$\pm 1.1$        & \textbf{86.5}$\mathbf{\pm 1.6}$    & 85.9$\pm 4.0$        & 83.0$\pm 2.6$        & \textbf{85.9}$\mathbf{\pm 2.5}$       & 85.5$\pm 0.8$        \\
Graph-SST2                & 89.7$\pm 0.5$         &  \textbf{89.9}$\mathbf{\pm 2.4}$       &   89.0$\pm 3.0$       &  92.2$\pm 0.3$        &92.0$\pm 0.2$         & \textbf{92.3}$\mathbf{\pm 0.4}$       & 88.1$\pm 0.8$          & \textbf{89.1}$\mathbf{\pm 1.2}$        &   88.7$\pm 0.9$       \\
Graph-Twitter             &   67.5$\pm 1.9$       &   \textbf{68.9}$\mathbf{\pm 5.9}$       &   66.1$\pm 6.5$        & 66.2$\pm 1.3$         & 75.2$\pm 2.8$        &    \textbf{76.5} $\mathbf{\pm 3.4}$      &\textbf{69.6}$\mathbf{\pm 6.5}$          & 64.8$\pm 4.0$        & 66.4$\pm 3.3$   \\
BA-Shape             &   91.9$\pm 1.7$       &   \textbf{95.7}$\mathbf{\pm 1.4}$       &   94.3$\pm 3.7$        & 92.9$\pm 0.5$         & 95.2$\pm 1.3$        &    \textbf{95.5} $\mathbf{\pm 2.4}$      &92.9$\pm 1.2$          & \textbf{93.4}$\mathbf{\pm 3.4}$        & 93.2$\pm 2.0$   \\
\bottomrule
\end{tabular}
\label{accuracy}
\end{table*}
\textbf{Datasets:} We conduct extensive experiments on different datasets and
GNN models to demonstrate the effectiveness of our proposed model. These datasets are listed as below:
\begin{itemize}
\item MUTAG \cite{debnath1991structure} and BBBP \cite{wu2018moleculenet} are molecule datasets for graph classification. In these datasets, nodes represent atoms and edges denote chemical bonds. The labels of molecular graphs are determined by the molecular compositions.
\item Graph-SST2 \cite{socher2013recursive} and Graph-Twitter \cite{dong2014adaptive} are sentiment graph datasets for graph classification. They convert sentences to graphs with Biaffine parser \cite{gardner2018allennlp}  that nodes denote words and edges represent the relationships
between words. The node embeddings are initialized with Bert  word embeddings \cite{devlin2018bert}. The labels are determined by the sentiment of text sentences.
\item BA-Shape is a synthetic node classification dataset. Each graph contains a base graph obtained from the Barab{\'a}si-Albert (BA) mode \cite{albert2002statistical} and a house-like five-node motif attached to the base graph. Each
node is labeled based on whether it belongs to the base graph or the different spatial locations of the motif.
\end{itemize}
\textbf{Experimental Settings:} In our evaluation, we use three variants of GNNs, i.e. GCN, GAT, and GIN. The split for train/validation/test sets is $80\%:10\%:10\%$. All models are trained for 500 epochs with an early stopping strategy based on accuracy on the validation set. We adopt the ADAM optimizer with a learning rate of 0.005. In Eq.(\ref{loss1}), the hyper-parameters $\lambda_1$, $\lambda_2$, and $\lambda_3$ are set to 0.10, 0.05, and 0.01 respectively. $s_{max}$ is set to 0.3 in Eq. (\ref{diversity loss}).  The number of prototypes per class $m$ is set to 5. In MCTS for prototype projection, we set $\lambda$ in Eq. (\ref{lambda}) to 5 and the number of iterations to 20. Each node in the Monte Carlo Tree can expand up to 10 child nodes and $N_{\rm min}$ is set to 5. The prototype  projection period $\tau$ is set to 50 and the projection epoch $T_p$ is set to 100. In the training of ProtGNN+, the warm-up epoch $T_w$ is set to 200. We employ a three-layer neural network to learn edge weights. In Eq. (\ref{loss2}), $\lambda_b$ is set to 0.01 and $B$ is set to 10. We select hyper-parameters based on related works or grid search, an analysis on hyper-parameters is included in the appendix.
All our experiments are conducted with one Tesla V100 GPU.
\subsection{Evaluations on ProtGNN/ProtGNN+}
\subsubsection{Comparison with Baselines}
In Table \ref{accuracy}, we compare the classification accuracy of ProtGNN/ProtGNN+ with the original GNNs. We apply 3 independent runs on random data splitting and report the means and standard deviations. In the following sections, we use GCN as the default backbone model. As we can see, ProtGNN and ProtGNN+ achieve comparable classification performance with the corresponding original GNN models, which also empirically verifies Theorem 1.

\begin{figure}[t]
	\centering
	\includegraphics[width=0.25\textwidth]{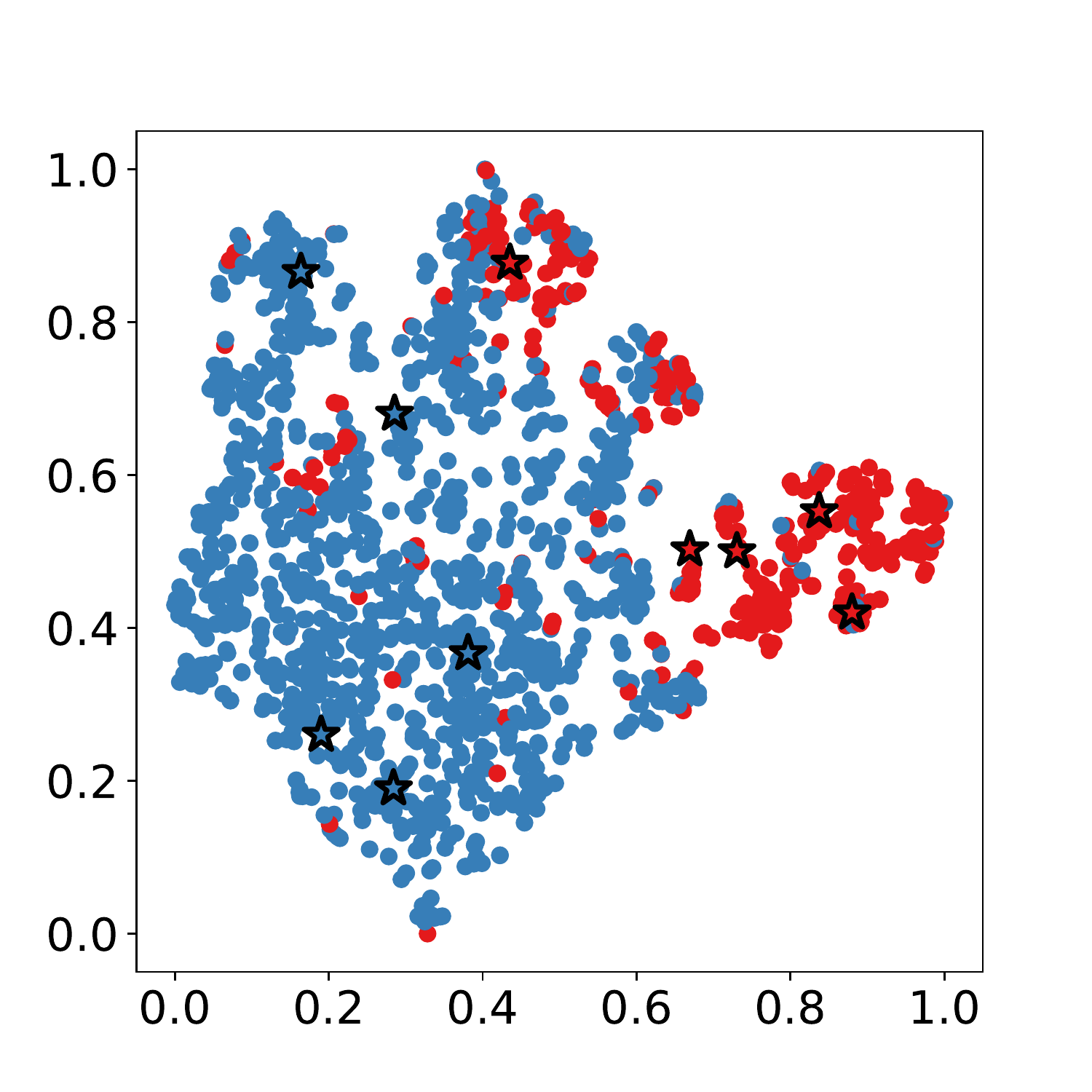}
	\caption{Visualization on BBBP dataset of the graph (dots) and prototype (stars) embeddings using the t-SNE method. Different colors indicate different classes.}
	\label{tsne}
\end{figure}

\subsubsection{Reasoning Process of Our Network}
In Figure \ref{case study 1}, we perform case studies on MUTAG and Graph-SST2 to qualitatively evaluate the performance of our proposed method. We visualize the prototypes and show the reasoning process of our ProtGNN+ in reaching a classification decision on input graphs. In particular, given an input graph $x$, the network finds the likelihood to be in each class by comparing it with prototypes from each class. The conditional subgraph sampling module finds the most similar subgraphs in $x$. These similarity scores are calculated, weighted, and summed together to give a final score for $x$ belonging to each class. For example, Figure \ref{case study 1}(a) shows the reasoning process of ProtGNN+ in deciding whether the input molecular graph is mutagenic. Based on chemical domain knowledge \cite{debnath1991structure}, carbon rings and $NO_2$ groups tend to be mutagenic. In the Prototype column of the mutagenic class, we can observe that the prototypes can capture the structures of $NO_2$ and carbon rings well. Moreover, in the column of Similar Subgraphs, the conditional subgraph sampling module can effectively identify the most similar subgraphs. For instance, in the first row of the mutagenic class, the $NO_2$ group and part of the carbon ring can be identified, which is quite similar to the prototype on the right.

Compared with biochemistry datasets, examples on text data could be more understandable since no special domain knowledge is required. In Figure \ref{case study 1}(b), the input graph ``can take the grandkids or the grandparents and never worry about anyone being bored" is positive. Our method can effectively capture the key phrase/subgraph leading to positiveness, ``never worry about bored". Furthermore, we can observe that the similarity score between the input graph with the positive prototypes e.g., ``kind of entertainment love to have" is much larger than negative prototypes e.g., ``embarrassed by invention". 

Overall, our method provides interpretable evidence to support classifications. Such explanations participate in the actual model computation and is always faithful to the classification decisions. More examples and case studies are reported in appendix.

\subsubsection{t-SNE Visualization of Prototypes}
In Figure \ref{tsne} we show the visualization on BBBP dataset of the graph and prototype embeddings using t-SNE method. We can observe that the prototypes can occupy the centers of graph embeddings, which verifies the effectiveness of prototype learning.

\begin{table}[t]
  \caption{Efficiency studies of different methods on BBBP}
\centering
\small
\begin{tabular}{ccccc}
\toprule
 Methods & GCN & ProtGNN & ProtGNN+ & ProtGNN+*  \\ \midrule
Time    & 177.9 s       & 506.3 s       & 632.7 s   & $\textgreater$ 2 hrs   \\
\bottomrule
\end{tabular}
\label{efficiency}
\end{table}
\subsubsection{Efficiency Studies} 
Finally, we study the efficiency of our proposed methods. In Table \ref{efficiency}, we show the time required to finish training for each model. Here ProtGNN+* denotes using MCTS for subgraph sampling in the training of ProtGNN+. The time complexity of ProtGNN+* is extremely high due to the complexity of MCTS. The proposed conditional subgraph sampling module can effectively reduce the time cost of ProtGNN+. Although ProtGNN and ProtGNN+ have a larger time cost compared to GCN (largely due to prototype projection with MCTS), the time cost is still acceptable considering the provided built-in interpretability. 
\section{Conclusion}
While extensive efforts have been made to explain GNNs from different angles, none of existing methods can provide \emph{built-in} explanations for GNNs. In this paper, we propose ProtGNN/ProtGNN+ which provides a new perspective on the explanations of GNNs. The prediction of ProtGNN is obtained by comparing the inputs to a few learned prototypes in the prototype layer. For better interpretability and higher efficiency, a novel conditional subgraph sampling module is proposed to indicate the subgraphs most similar to prototypes. Extensive experimental results show that our method can provide a human-intelligible reasoning process with acceptable classification accuracy and time-complexity. 

\bibliography{aaai22}

\begin{thebibliography}{35}
\providecommand{\natexlab}[1]{#1}

\bibitem[{Albert and Barab{\'a}si(2002)}]{albert2002statistical}
Albert, R.; and Barab{\'a}si, A.-L. 2002.
\newblock Statistical mechanics of complex networks.
\newblock \emph{Reviews of modern physics}, 74(1): 47.

\bibitem[{Baldassarre and Azizpour(2019)}]{baldassarre2019explainability}
Baldassarre, F.; and Azizpour, H. 2019.
\newblock Explainability techniques for graph convolutional networks.
\newblock \emph{ICML workshop}.

\bibitem[{Bian et~al.(2020)Bian, Xiao, Xu, Zhao, Huang, Rong, and
  Huang}]{bian2020rumor}
Bian, T.; Xiao, X.; Xu, T.; Zhao, P.; Huang, W.; Rong, Y.; and Huang, J. 2020.
\newblock Rumor detection on social media with bi-directional graph
  convolutional networks.
\newblock In \emph{AAAI}, volume~34, 549--556.

\bibitem[{Chen et~al.(2018)Chen, Li, Tao, Barnett, Su, and
  Rudin}]{chen2018looks}
Chen, C.; Li, O.; Tao, C.; Barnett, A.~J.; Su, J.; and Rudin, C. 2018.
\newblock This looks like that: deep learning for interpretable image
  recognition.
\newblock \emph{NeurIPS}.

\bibitem[{Debnath et~al.(1991)Debnath, Lopez~de Compadre, Debnath, Shusterman,
  and Hansch}]{debnath1991structure}
Debnath, A.~K.; Lopez~de Compadre, R.~L.; Debnath, G.; Shusterman, A.~J.; and
  Hansch, C. 1991.
\newblock Structure-activity relationship of mutagenic aromatic and
  heteroaromatic nitro compounds. correlation with molecular orbital energies
  and hydrophobicity.
\newblock \emph{Journal of medicinal chemistry}, 34(2): 786--797.

\bibitem[{Devlin et~al.(2018)Devlin, Chang, Lee, and
  Toutanova}]{devlin2018bert}
Devlin, J.; Chang, M.-W.; Lee, K.; and Toutanova, K. 2018.
\newblock Bert: Pre-training of deep bidirectional transformers for language
  understanding.
\newblock \emph{arXiv preprint arXiv:1810.04805}.

\bibitem[{Dong et~al.(2014)Dong, Wei, Tan, Tang, Zhou, and
  Xu}]{dong2014adaptive}
Dong, L.; Wei, F.; Tan, C.; Tang, D.; Zhou, M.; and Xu, K. 2014.
\newblock Adaptive recursive neural network for target-dependent twitter
  sentiment classification.
\newblock In \emph{ACL}, 49--54.

\bibitem[{Gardner et~al.(2018)Gardner, Grus, Neumann, Tafjord, Dasigi, Liu,
  Peters, Schmitz, and Zettlemoyer}]{gardner2018allennlp}
Gardner, M.; Grus, J.; Neumann, M.; Tafjord, O.; Dasigi, P.; Liu, N.; Peters,
  M.; Schmitz, M.; and Zettlemoyer, L. 2018.
\newblock Allennlp: A deep semantic natural language processing platform.
\newblock \emph{arXiv preprint arXiv:1803.07640}.

\bibitem[{Gilbert(1959)}]{gilbert1959random}
Gilbert, E.~N. 1959.
\newblock Random graphs.
\newblock \emph{The Annals of Mathematical Statistics}, 30(4): 1141--1144.

\bibitem[{Gilmer et~al.(2017)Gilmer, Schoenholz, Riley, Vinyals, and
  Dahl}]{gilmer2017neural}
Gilmer, J.; Schoenholz, S.~S.; Riley, P.~F.; Vinyals, O.; and Dahl, G.~E. 2017.
\newblock Neural message passing for quantum chemistry.
\newblock In \emph{ICML}, 1263--1272. PMLR.

\bibitem[{He et~al.(2010)He, Zhang, Shi, Hu, Kong, Cai, and
  Chou}]{he2010predicting}
He, Z.; Zhang, J.; Shi, X.-H.; Hu, L.-L.; Kong, X.; Cai, Y.-D.; and Chou, K.-C.
  2010.
\newblock Predicting drug-target interaction networks based on functional
  groups and biological features.
\newblock \emph{PloS one}, 5(3): e9603.

\bibitem[{Huang et~al.(2020)Huang, Yamada, Tian, Singh, Yin, and
  Chang}]{huang2020graphlime}
Huang, Q.; Yamada, M.; Tian, Y.; Singh, D.; Yin, D.; and Chang, Y. 2020.
\newblock Graphlime: Local interpretable model explanations for graph neural
  networks.
\newblock \emph{arXiv preprint arXiv:2001.06216}.

\bibitem[{Kipf and Welling(2017)}]{kipf2016semi}
Kipf, T.~N.; and Welling, M. 2017.
\newblock Semi-supervised classification with graph convolutional networks.
\newblock \emph{ICLR}.

\bibitem[{Kolodner(1992)}]{kolodner1992introduction}
Kolodner, J.~L. 1992.
\newblock An introduction to case-based reasoning.
\newblock \emph{Artificial intelligence review}, 6(1): 3--34.

\bibitem[{Luo et~al.(2020)Luo, Cheng, Xu, Yu, Zong, Chen, and
  Zhang}]{luo2020parameterized}
Luo, D.; Cheng, W.; Xu, D.; Yu, W.; Zong, B.; Chen, H.; and Zhang, X. 2020.
\newblock Parameterized explainer for graph neural network.
\newblock \emph{NeurIPS}.

\bibitem[{Ming et~al.(2019)Ming, Xu, Qu, and Ren}]{ming2019interpretable}
Ming, Y.; Xu, P.; Qu, H.; and Ren, L. 2019.
\newblock Interpretable and steerable sequence learning via prototypes.
\newblock In \emph{SIGKDD}.

\bibitem[{Pope et~al.(2019)Pope, Kolouri, Rostami, Martin, and
  Hoffmann}]{pope2019explainability}
Pope, P.~E.; Kolouri, S.; Rostami, M.; Martin, C.~E.; and Hoffmann, H. 2019.
\newblock Explainability methods for graph convolutional neural networks.
\newblock In \emph{CVPR}, 10772--10781.

\bibitem[{Rudin(2018)}]{rudin2018please}
Rudin, C. 2018.
\newblock Please stop explaining black box models for high stakes decisions.
\newblock \emph{stat}, 1050: 26.

\bibitem[{Rymarczyk et~al.(2021)Rymarczyk, Struski, Tabor, and
  Zieli{\'n}ski}]{rymarczyk2020protopshare}
Rymarczyk, D.; Struski, {\L}.; Tabor, J.; and Zieli{\'n}ski, B. 2021.
\newblock ProtoPShare: Prototype Sharing for Interpretable Image Classification
  and Similarity Discovery.
\newblock \emph{SIGKDD}.

\bibitem[{Schlichtkrull, De~Cao, and
  Titov(2020)}]{schlichtkrull2020interpreting}
Schlichtkrull, M.~S.; De~Cao, N.; and Titov, I. 2020.
\newblock Interpreting graph neural networks for nlp with differentiable edge
  masking.
\newblock \emph{arXiv preprint arXiv:2010.00577}.

\bibitem[{Schmidt et~al.(2001)Schmidt, Montani, Bellazzi, Portinale, and
  Gierl}]{schmidt2001cased}
Schmidt, R.; Montani, S.; Bellazzi, R.; Portinale, L.; and Gierl, L. 2001.
\newblock Cased-based reasoning for medical knowledge-based systems.
\newblock \emph{International Journal of Medical Informatics}, 64(2-3):
  355--367.

\bibitem[{Schnake et~al.(2020)Schnake, Eberle, Lederer, Nakajima, Sch{\"u}tt,
  M{\"u}ller, and Montavon}]{schnake2020xai}
Schnake, T.; Eberle, O.; Lederer, J.; Nakajima, S.; Sch{\"u}tt, K.~T.;
  M{\"u}ller, K.-R.; and Montavon, G. 2020.
\newblock XAI for graphs: explaining graph neural network predictions by
  identifying relevant walks.
\newblock \emph{arXiv e-prints}, arXiv--2006.

\bibitem[{Schwarzenberg et~al.(2019)Schwarzenberg, H{\"u}bner, Harbecke, Alt,
  and Hennig}]{schwarzenberg2019layerwise}
Schwarzenberg, R.; H{\"u}bner, M.; Harbecke, D.; Alt, C.; and Hennig, L. 2019.
\newblock Layerwise relevance visualization in convolutional text graph
  classifiers.
\newblock \emph{arXiv preprint arXiv:1909.10911}.

\bibitem[{Silver et~al.(2017)Silver, Schrittwieser, Simonyan, Antonoglou,
  Huang, Guez, Hubert, Baker, Lai, Bolton et~al.}]{silver2017mastering}
Silver, D.; Schrittwieser, J.; Simonyan, K.; Antonoglou, I.; Huang, A.; Guez,
  A.; Hubert, T.; Baker, L.; Lai, M.; Bolton, A.; et~al. 2017.
\newblock Mastering the game of go without human knowledge.
\newblock \emph{nature}, 550(7676): 354--359.

\bibitem[{Socher et~al.(2013)Socher, Perelygin, Wu, Chuang, Manning, Ng, and
  Potts}]{socher2013recursive}
Socher, R.; Perelygin, A.; Wu, J.; Chuang, J.; Manning, C.~D.; Ng, A.~Y.; and
  Potts, C. 2013.
\newblock Recursive deep models for semantic compositionality over a sentiment
  treebank.
\newblock In \emph{EMNLP}, 1631--1642.

\bibitem[{Veli{\v{c}}kovi{\'c} et~al.(2017)Veli{\v{c}}kovi{\'c}, Cucurull,
  Casanova, Romero, Lio, and Bengio}]{velivckovic2017graph}
Veli{\v{c}}kovi{\'c}, P.; Cucurull, G.; Casanova, A.; Romero, A.; Lio, P.; and
  Bengio, Y. 2017.
\newblock Graph attention networks.
\newblock \emph{arXiv preprint arXiv:1710.10903}.

\bibitem[{Vu and Thai(2020)}]{NEURIPS2020_8fb134f2}
Vu, M.; and Thai, M.~T. 2020.
\newblock PGM-Explainer: Probabilistic Graphical Model Explanations for Graph
  Neural Networks.
\newblock In Larochelle, H.; Ranzato, M.; Hadsell, R.; Balcan, M.~F.; and Lin,
  H., eds., \emph{NeurIPS}, volume~33.

\bibitem[{Wu et~al.(2018)Wu, Ramsundar, Feinberg, Gomes, Geniesse, Pappu,
  Leswing, and Pande}]{wu2018moleculenet}
Wu, Z.; Ramsundar, B.; Feinberg, E.~N.; Gomes, J.; Geniesse, C.; Pappu, A.~S.;
  Leswing, K.; and Pande, V. 2018.
\newblock MoleculeNet: a benchmark for molecular machine learning.
\newblock \emph{Chemical science}, 9(2): 513--530.

\bibitem[{Xu et~al.(2019)Xu, Hu, Leskovec, and Jegelka}]{xu2018powerful}
Xu, K.; Hu, W.; Leskovec, J.; and Jegelka, S. 2019.
\newblock How Powerful are Graph Neural Networks?
\newblock In \emph{ICLR}.

\bibitem[{Yang et~al.(2020)Yang, Zhang, Zhou, Wang, Sun, Zhong, Fang, Yu, and
  Qi}]{yang2020financial}
Yang, S.; Zhang, Z.; Zhou, J.; Wang, Y.; Sun, W.; Zhong, X.; Fang, Y.; Yu, Q.;
  and Qi, Y. 2020.
\newblock Financial Risk Analysis for SMEs with Graph-based Supply Chain
  Mining.
\newblock In \emph{IJCAI}, 4661--4667.

\bibitem[{Ying et~al.(2019)Ying, Bourgeois, You, Zitnik, and
  Leskovec}]{ying2019gnnexplainer}
Ying, R.; Bourgeois, D.; You, J.; Zitnik, M.; and Leskovec, J. 2019.
\newblock Gnnexplainer: Generating explanations for graph neural networks.
\newblock \emph{NeurIPS}, 32: 9240.

\bibitem[{Yuan et~al.(2020{\natexlab{a}})Yuan, Tang, Hu, and Ji}]{yuan2020xgnn}
Yuan, H.; Tang, J.; Hu, X.; and Ji, S. 2020{\natexlab{a}}.
\newblock Xgnn: Towards model-level explanations of graph neural networks.
\newblock In \emph{SIGKDD}, 430--438.

\bibitem[{Yuan et~al.(2020{\natexlab{b}})Yuan, Yu, Gui, and
  Ji}]{yuan2020explainability}
Yuan, H.; Yu, H.; Gui, S.; and Ji, S. 2020{\natexlab{b}}.
\newblock Explainability in graph neural networks: A taxonomic survey.
\newblock \emph{arXiv preprint arXiv:2012.15445}.

\bibitem[{Yuan et~al.(2021)Yuan, Yu, Wang, Li, and Ji}]{yuan2021explainability}
Yuan, H.; Yu, H.; Wang, J.; Li, K.; and Ji, S. 2021.
\newblock On explainability of graph neural networks via subgraph explorations.
\newblock \emph{ICML}.

\bibitem[{Zhang et~al.(2021)Zhang, Liu, Wang, Lu, and Lee}]{zhang2021motif}
Zhang, Z.; Liu, Q.; Wang, H.; Lu, C.; and Lee, C.-K. 2021.
\newblock Motif-based Graph Self-Supervised Learning for Molecular Property
  Prediction.
\newblock \emph{NeurIPS}, 34.

\end{thebibliography}

\appendix

\clearpage

\begin{table*}[t]
  \caption{Statistics of five datasets}
\centering
\begin{tabular}{cccccc}
\toprule
\multirow{2}{*}{} & \multicolumn{5}{c}{Datasets}              \\ \cmidrule(r){2-6} 
                          & MUTAG & BBBP & Graph-SST2 & Graph-Twitter & BA-Shape  \\ \midrule
$\#$ of Edges (avg)                    & 19.79       & 25.95       & 9.20        & 20.10        & 2055        \\
$\#$ of Nodes (avg)                    & 17.93        & 24.06      & 10.19        & 21.10        & 700       \\
$\#$ of Graphs               & 188         &  2039       & 70042      & 6940  &   1      \\
$\#$ of Classes             &   2      &   2       &   2        &  3        &   4     \\
\bottomrule
\end{tabular}
\label{statistics}
\end{table*}
\section{Dataset Statistics}
In Table \ref{statistics}, we show the detailed statistics of five datasets. These datasets include biological data, text data, and synthetic data. The first four datasets are used for graph classification tasks while BA-Shape is used for node classfication.

\begin{figure*}[t]
	\centering
	\includegraphics[width=0.9\textwidth]{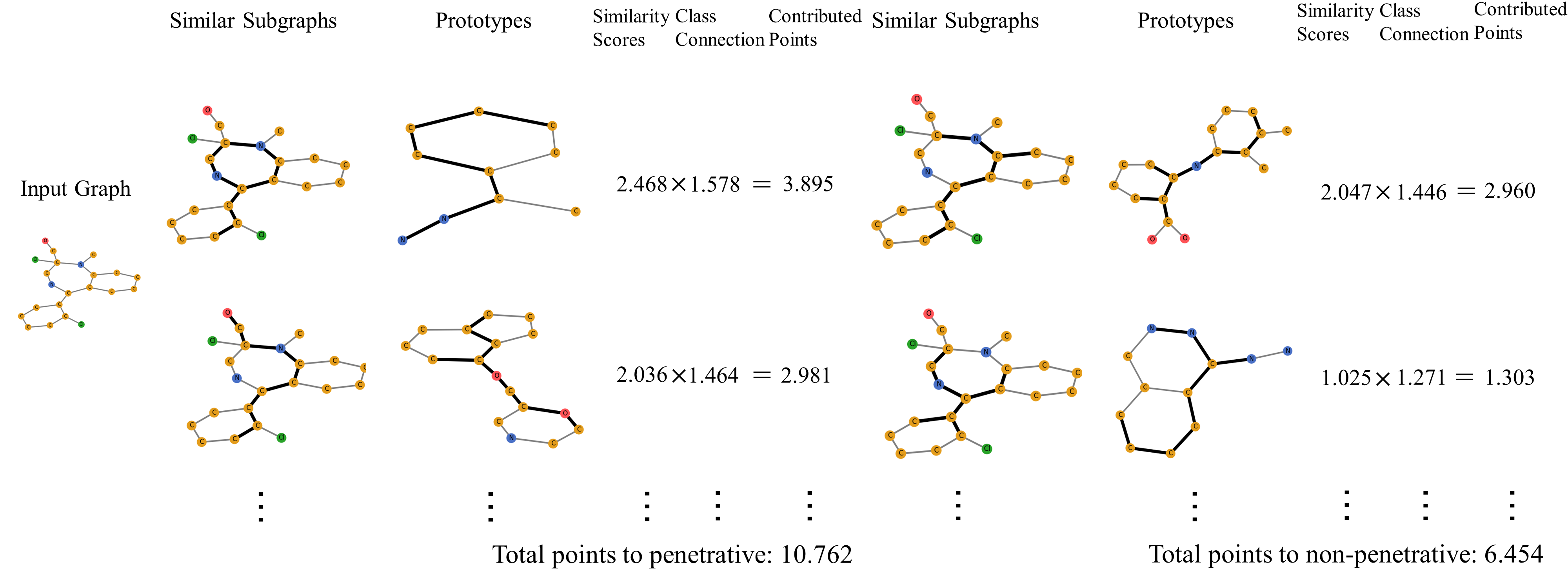}
	\caption{The reasoning process of ProtGNN+ in deciding whether the input molecular graph is penetrative.}
	\label{case study 2}
\end{figure*}

\begin{figure*}[t]
	\centering
	\subfigure[]{

\begin{minipage}[b]{0.75\textwidth}
\includegraphics[width=1\textwidth]{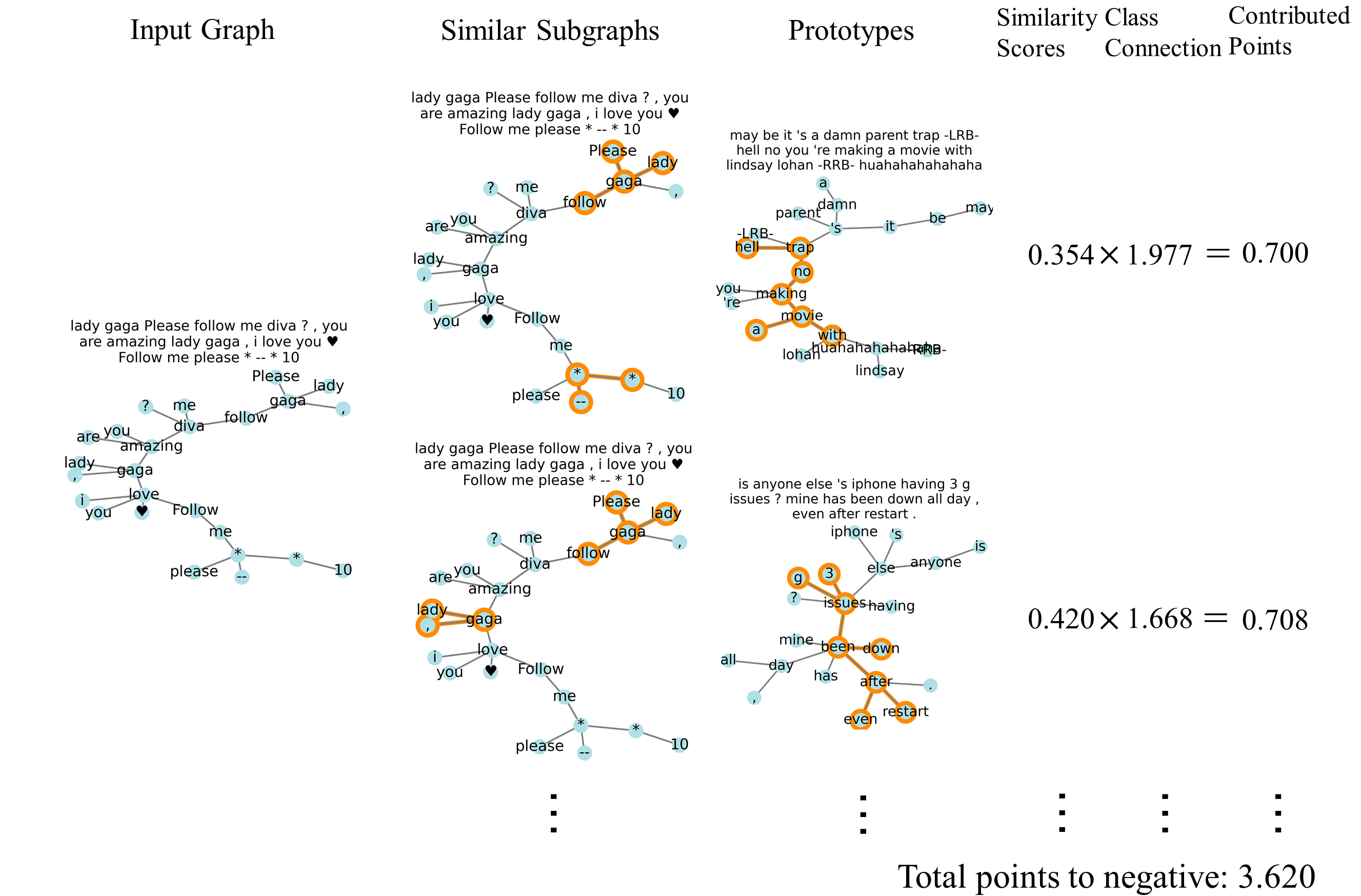}
\end{minipage}
}

\subfigure[]{
\begin{minipage}[b]{0.95\textwidth}
\includegraphics[width=1\textwidth]{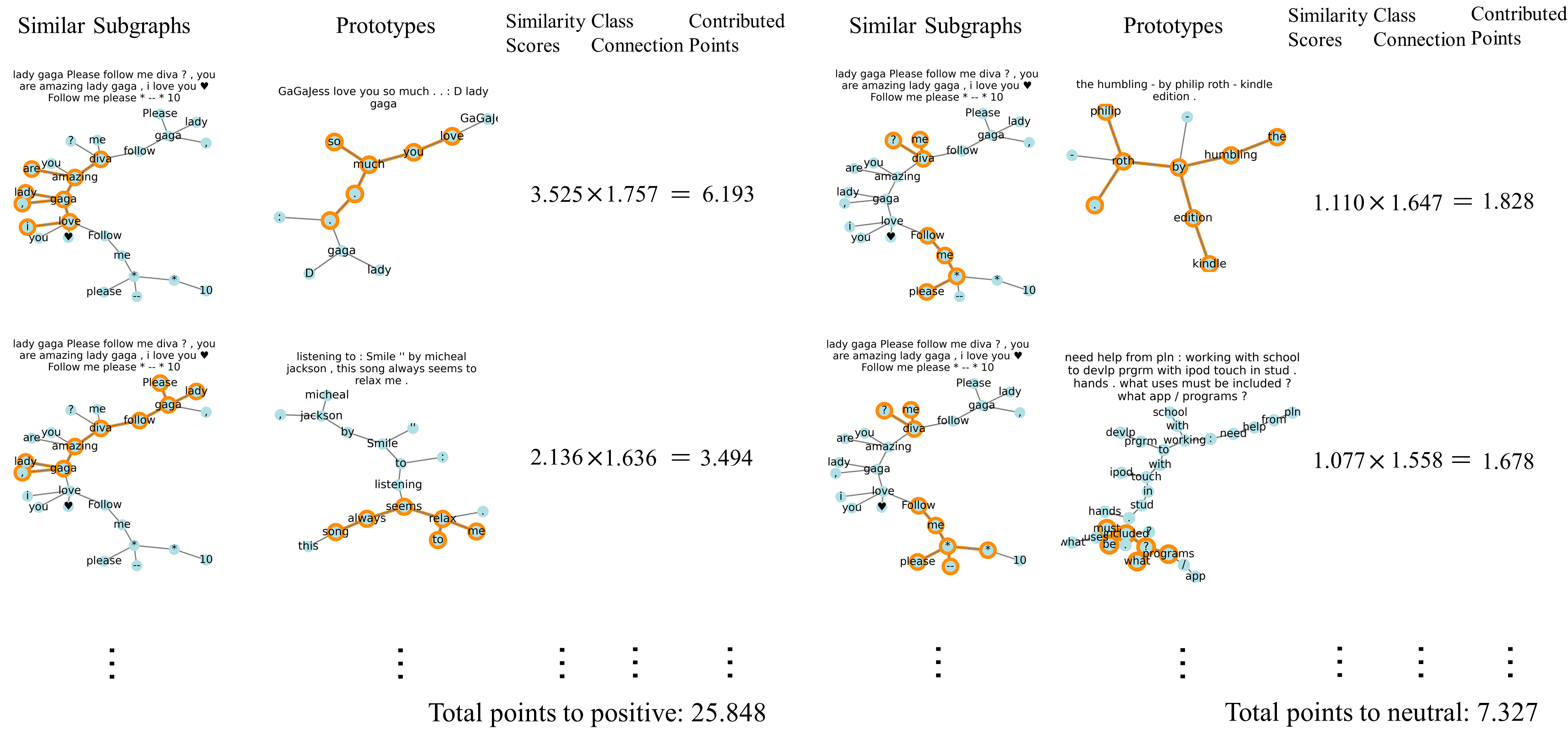}
\end{minipage}
}
\caption{The reasoning process of ProtGNN+ in deciding whether the sentiment of the text graph is positive, neutral, or negative.}

\label{case study 3}
\end{figure*}

\section{Proof of Theorem 1}
In this section, we provide a proof for Theorem 1 in the main paper.\\
\textbf{Theorem 1:} \label{theorem}Let $c\circ g_p \circ f$ be a ProtoGNN. The embedding of the input graph is $h$. We assume that the number of prototypes is the same for each class, which is denoted as $m$. For each class $k$, the weight connection in the last layer $c$ between a class $k$ prototype and the class $k$ logit is 1, and that between a non-class $k$ prototype and the class $k$ logit is 0. Let $p_l^k$ denote the $l$-th prototype for class $k$ and  $h_l^k$ the embedding of the pruned subgraph. ProtGNN and ProtGNN+ has the same graph encoder $f$.\\
We make the following assumptions: there exists some $\delta$ with $0<\delta<1$,
\begin{itemize}
    \item for the correct class, we have $\|h-h_l^k\|_2 \leq (\sqrt{1+\delta}-1)\|h-p_l^k\|_2$ and $\|h-p_l^k\|_2 \leq \sqrt{1-\delta}$;
    \item for the incorrect classes, $\|h-h_l^k\|_2 \leq \theta\|h-p_l^k\|_2 - \sqrt{\epsilon}$, $\theta = {\rm min}(\sqrt{1+\delta}-1, 1-\frac{1}{\sqrt{2-\delta}})$.
\end{itemize}
For one correctly classified input graph in ProtGNN, if the output logits between the top-2 classes are at least $2m log((1+\delta)(2-\delta))$, then ProtGNN+ can classify the input graph correctly as well.

\textit{Proof:} For any class $k$, let $L_k(x, \{p_l^k\}_{l=1}^m)$ denotes the summed contributed scores for graph $x$ belonging to class $k$ in ProtGNN. According to Eq. (\ref{score function}) and the assumption:
\begin{equation}
    L_k(x, \{p_l^k\}_{l=1}^m)=\sum_{l=1}^m log(\frac{\|h-p_l^k\|_2^2 + 1}{\|h-p_l^k\|_2^2 + \epsilon}).
\end{equation}
Let $L'_k(x, \{p_l^k\}_{l=1}^m)$ denotes the summed contributed scores in ProtGNN+:
\begin{equation}
    L'_k(x, \{p_l^k\}_{l=1}^m)=\sum_{l=1}^m log(\frac{\|h_l^k-p_l^k\|_2^2 + 1}{\|h_l^k-p_l^k\|_2^2 + \epsilon}).
\end{equation}
Then, the gap between the summend contributed scores denoted by $\Delta_k$ is:
\begin{equation}
\begin{aligned}
    \Delta_k &= L'_k(x, \{p_l^k\}_{l=1}^m) - L_k(x, \{p_l^k\}_{l=1}^m)\\
    &= \sum_{l=1}^m log(\frac{\|h_l^k-p_l^k\|_2^2 + 1}{\|h-p_l^k\|_2^2 + 1} \cdot \frac{\|h-p_l^k\|_2^2 + \epsilon}{\|h_l^k-p_l^k\|_2^2 + \epsilon}).
\end{aligned}
\end{equation}
\textbf{Correct class:} We first derive the lower bound of $\Delta_k$ for the correct class. Firstly, we have
\begin{equation}
    \frac{\|h_l^k-p_l^k\|_2^2 + 1}{\|h-p_l^k\|_2^2 + 1} \ge \frac{1}{\|h-p_l^k\|_2^2 + 1} \ge \frac{1}{2-\delta}
\end{equation}
Then, by the triangle inequality , we have $\|h_l^k-p_l^k\| \le \|h-p_l^k\| + \|h-h_l^k\|$. As a result, we have:
\begin{equation}
\begin{aligned}
    \frac{\|h-p_l^k\|_2^2 + \epsilon}{\|h_l^k-p_l^k\|_2^2 + \epsilon} &\ge \frac{\|h-p_l^k\|_2^2 + \epsilon}{(\|h-p_l^k\|_2 + \|h-h_l^k\|_2)^2 + \epsilon}\\
    & \ge \frac{\|h-p_l^k\|_2^2 + \epsilon}{(1+\delta)\|h-p_l^k\|_2^2 + \epsilon}\\
    &\ge \frac{1}{1+\delta}
\end{aligned}
\end{equation}
Combining the above two inequalities, $\Delta_k$ for the correct class is $m {\rm log}(\frac{1}{(1+\delta)(2-\delta)}) = -m {\rm log}((1+\delta)(2-\delta))$.\\
\textbf{Wrong class:} Now we begin to derive an upper bound of $\Delta_k$ for incorrect classes. First,
\begin{equation}
\begin{aligned}
    \frac{\|h_l^k-p_l^k\|_2^2 + 1}{\|h-p_l^k\|_2^2 + 1} & \le \frac{(\|h-p_l^k\|_2+ \|h-h_l^k\|)^2 + 1}{\|h-p_l^k\|_2^2 + 1}\\
    &\le \frac{(\|h-p_l^k\|_2+ (\sqrt{1+\delta}-1)\|h-p_l^k\|_2)^2 + 1}{\|h-p_l^k\|_2^2 + 1}\\
    &\le 1+\delta
\end{aligned}
\end{equation}
Then,
\begin{equation}
    \begin{aligned}
    \frac{\|h-p_l^k\|_2^2 + \epsilon}{\|h_l^k-p_l^k\|_2^2 + \epsilon} &\le \frac{\|h-p_l^k\|_2^2 + \epsilon}{(\|h-p_l^k\|_2 - \|h-h_l^k\|_2)^2 + \epsilon}\\
    &\le (\frac{\|h-p_l^k\|_2 + \sqrt{\epsilon}}{\|h-p_l^k\|_2 - \|h-h_l^k\|_2})^2
    \end{aligned}
    \label{1}
\end{equation}
According to the assumption for incorrect classes, we have:
\begin{equation}
\|h-h_l^k\|_2 \le (1-\frac{1}{\sqrt{2-\delta}})\|h-p_l^k\|_2-\sqrt{\epsilon}
\end{equation}
\begin{equation}
    \frac{1}{\sqrt{2-\delta}}\|h-p_l^k\|_2 + \sqrt{\epsilon} \le \|h-p_l^k\|_2 - \|h-h_l^k\|_2
    \label{2}
\end{equation}
Combining Eq. (\ref{1}) and Eq. (\ref{2}), we have:
\begin{equation}
    \frac{\|h-p_l^k\|_2^2 + \epsilon}{\|h_l^k-p_l^k\|_2^2 + \epsilon} \le (\sqrt{2-\delta})^2=2-\delta
\end{equation}
For incorrect classes, $\Delta_k \ge m {\rm log}((1+\delta)(2-\delta))$.

Finally, suppose the summed contributed scores of the correct class is at least $2m {\rm log}((1+\delta)(2-\delta))$ larger than any other classes in ProtGNN, the input graph will still be correctly classified by ProtGNN+. $\hfill\qedsymbol$
\section{Architecture of the Conditional Subgraph Sampling Module}
\begin{table}[h]
\centering
\caption{Architecture of the Conditional Subgraph Sampling Module}
\begin{tabular}{|c|c|}

\hline
Layer                  & Size                 \\ \hline
Input                  & 128 + 128 + 128 \\ \hline
Fully Connected + ReLU    & 64  \\ \hline
Fully Connected + ReLU     & 8  \\ \hline
Sigmoid               & 1  \\ \hline
\end{tabular}
\label{elayer}
\end{table}
In the conditional subgraph sampling module, we adopt deep neural networks to learn $e_{ij}$: 
\begin{equation}
    e_{ij} = \sigma({\rm MLP}_{\theta} ([z_i;z_j;p_k])).
\end{equation}
In Table \ref{elayer}, we show the details of architecture. In our experiments, the node embedding size and prototype size are set to 128. To make sure the selected adjacency matrix is symmetric, we set $\mathcal{E}$ as $\frac{\mathcal{E}+\mathcal{E}^T}{2}$ in experiments.
\section{More Case Studies}
\begin{figure}[t]
	\centering
    \subfigure[BBBP]{\includegraphics[width=0.23\textwidth]{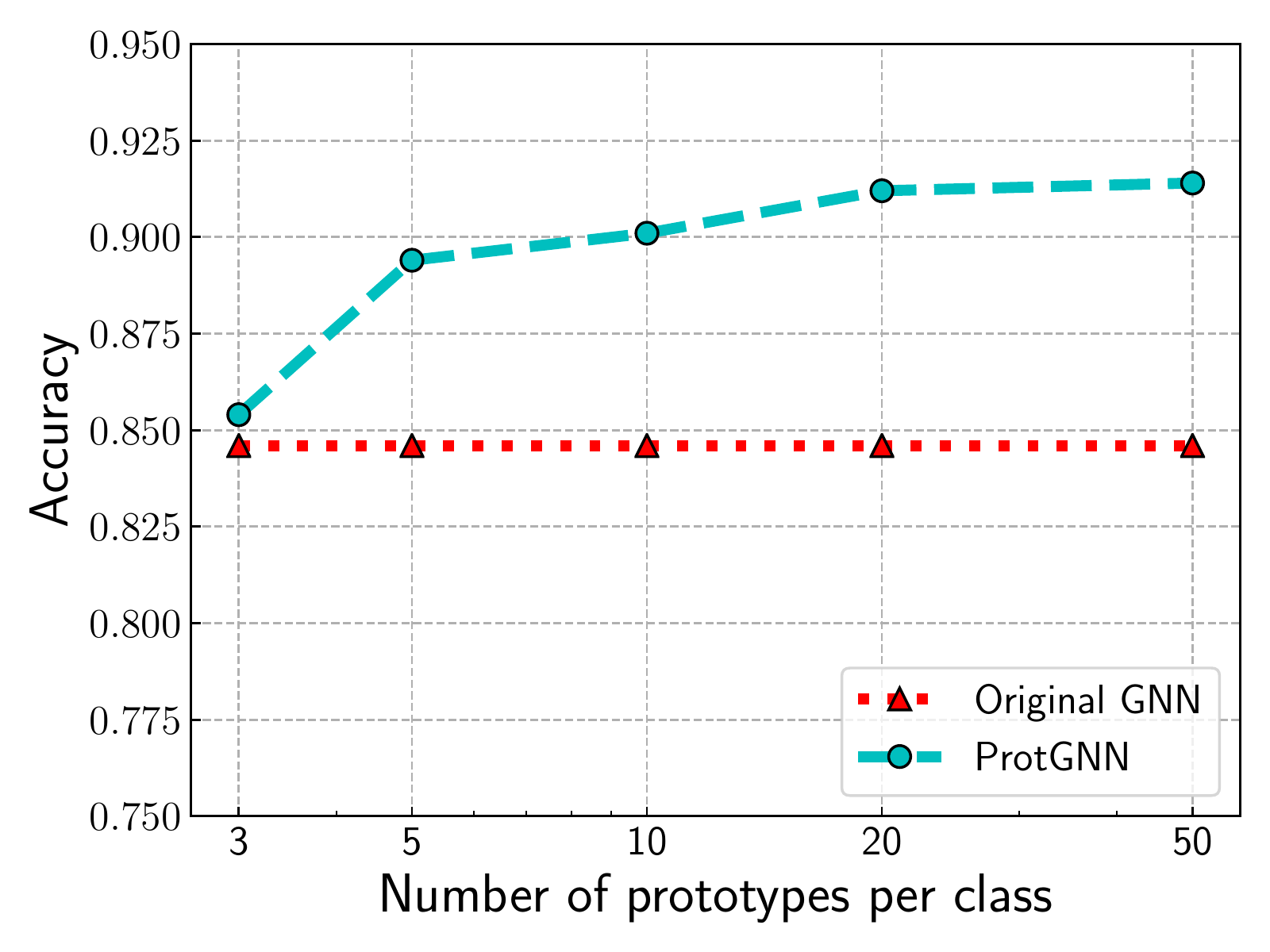}}
    \subfigure[Graph-Twitter]{\includegraphics[width=0.23\textwidth]{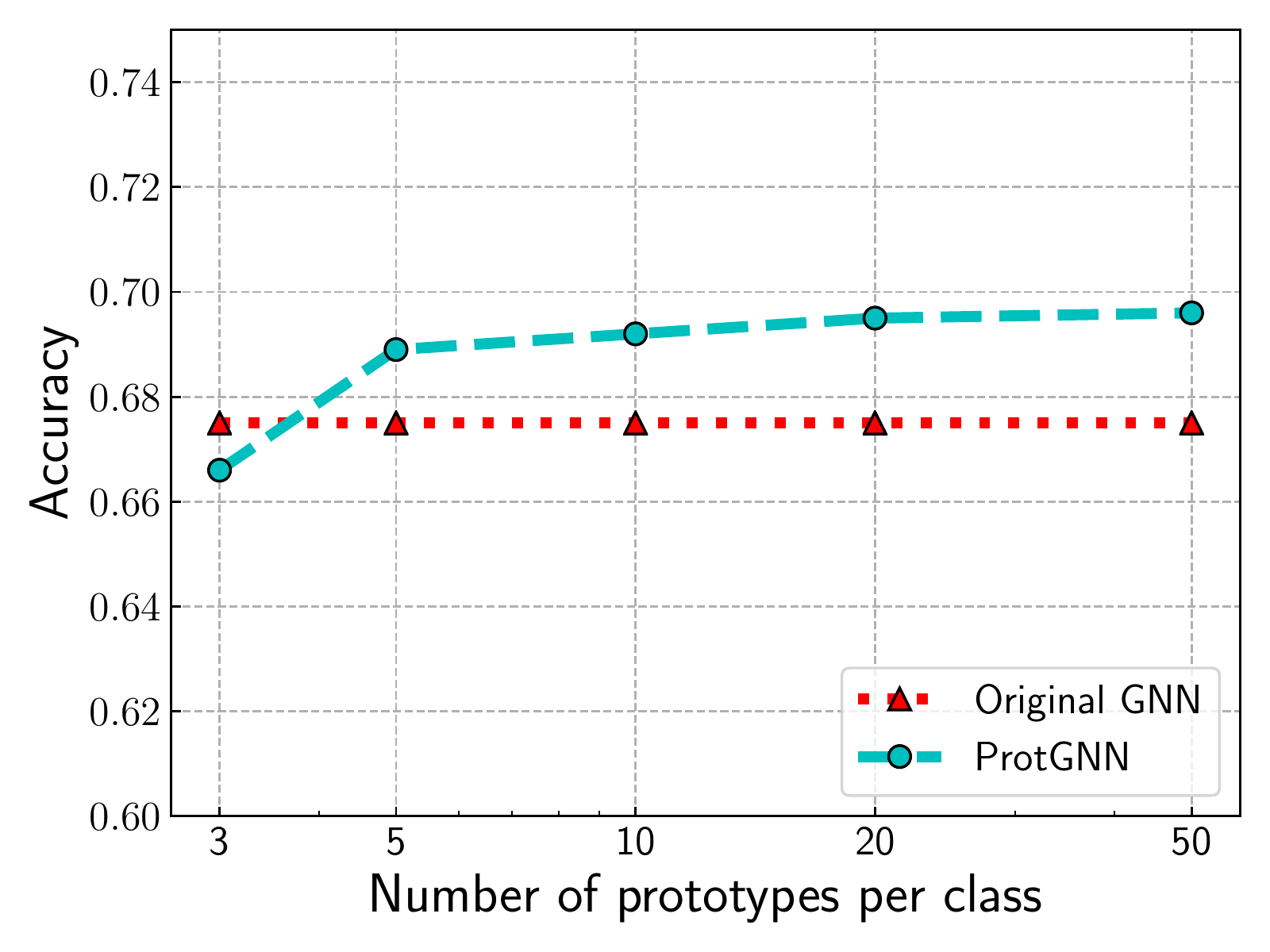}}
    \caption{The influence of the number of prototypes per class $m$ on performance. The accuracy of ProtGNN firstly increase dramatically as $m$ increases. Then the increasing slope flattens after $m$ exceeds 5.}
    \label{m}
\end{figure}

\begin{figure}[t]
	\centering
    \subfigure[w/ Div Class 0]{\includegraphics[width=0.23\textwidth]{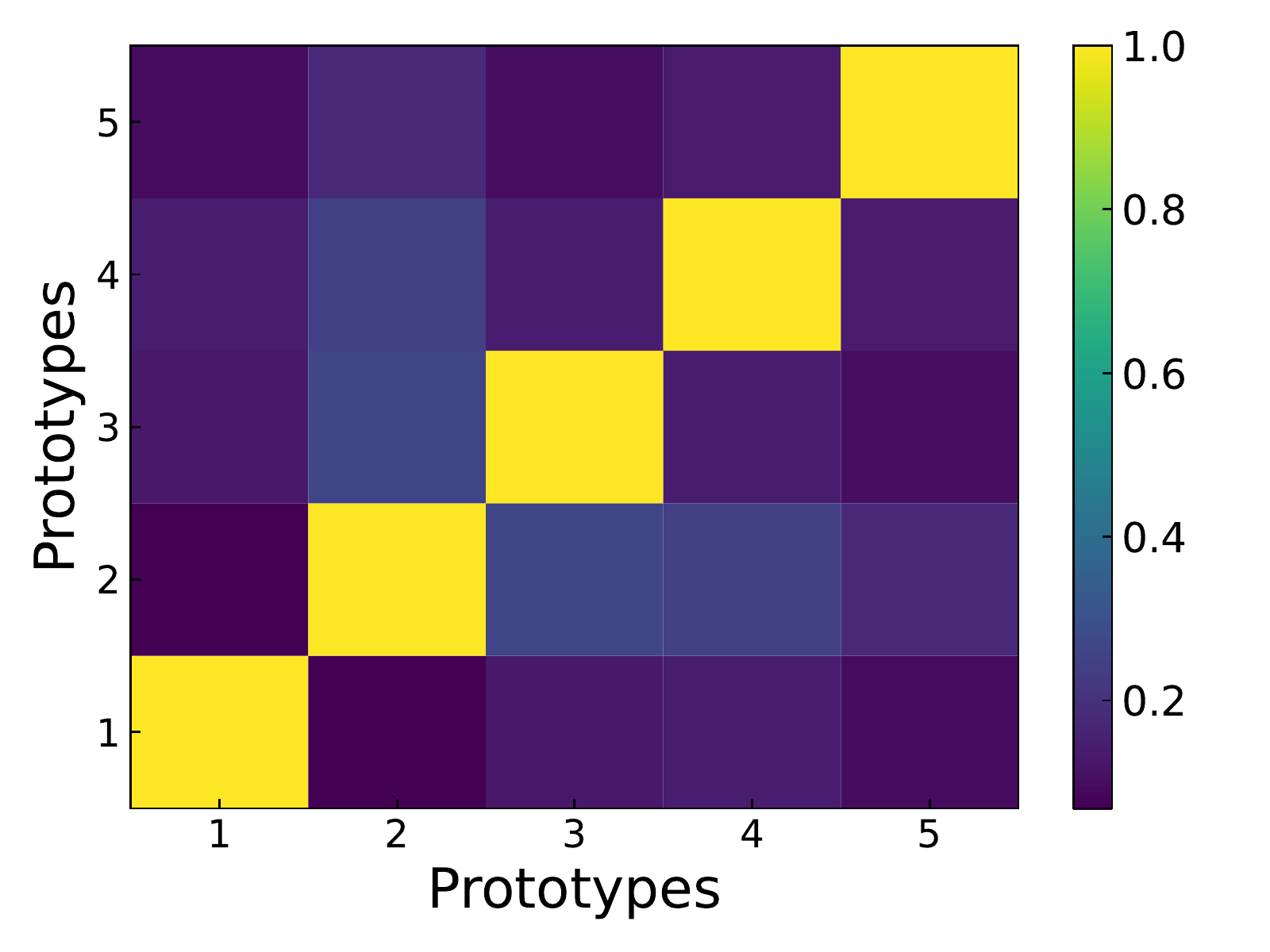}}
    \subfigure[w/ Div Class 1]{\includegraphics[width=0.23\textwidth]{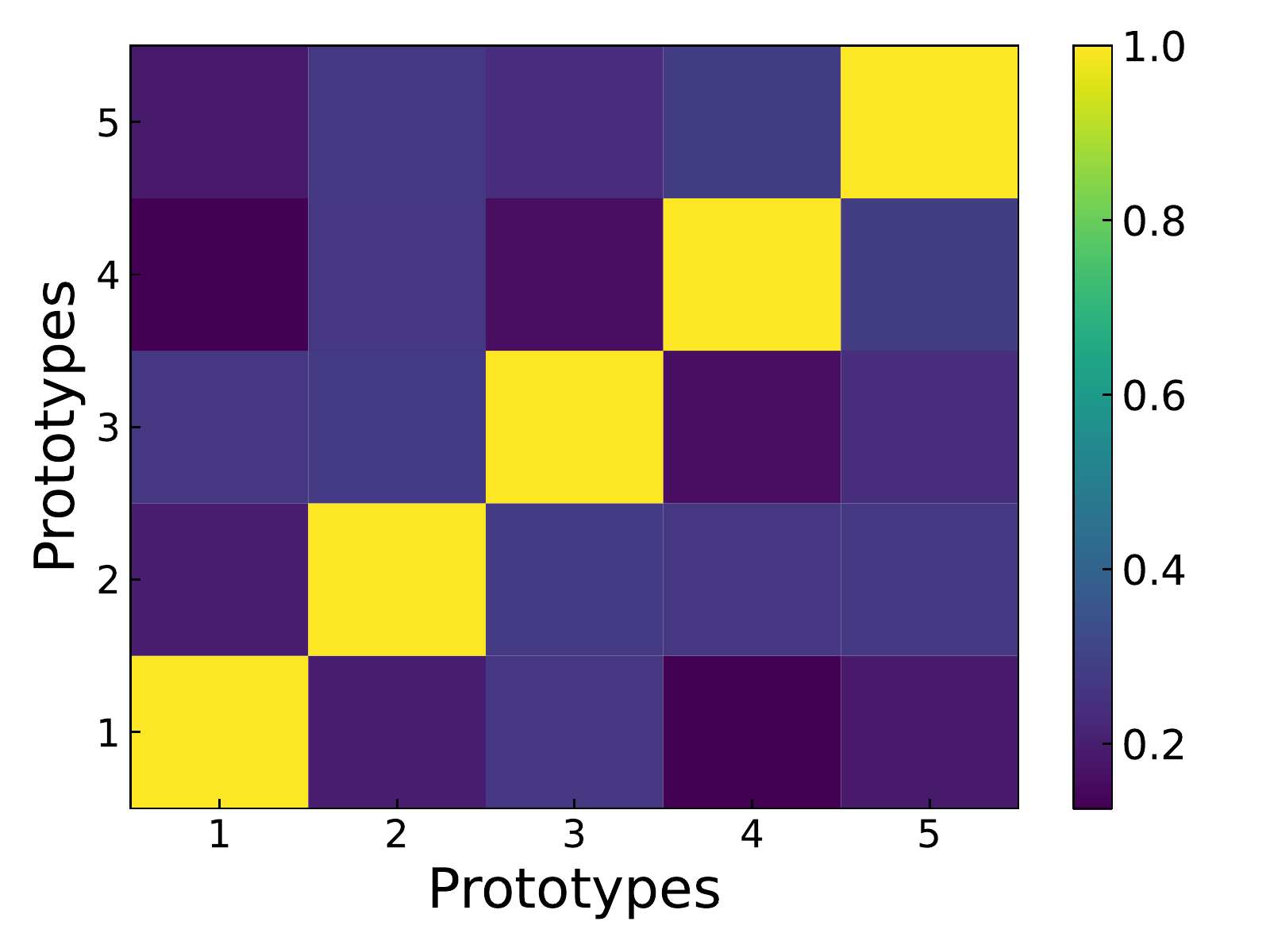}}
    \quad
    
    \subfigure[w/o Div Class 0]{\includegraphics[width=0.23\textwidth]{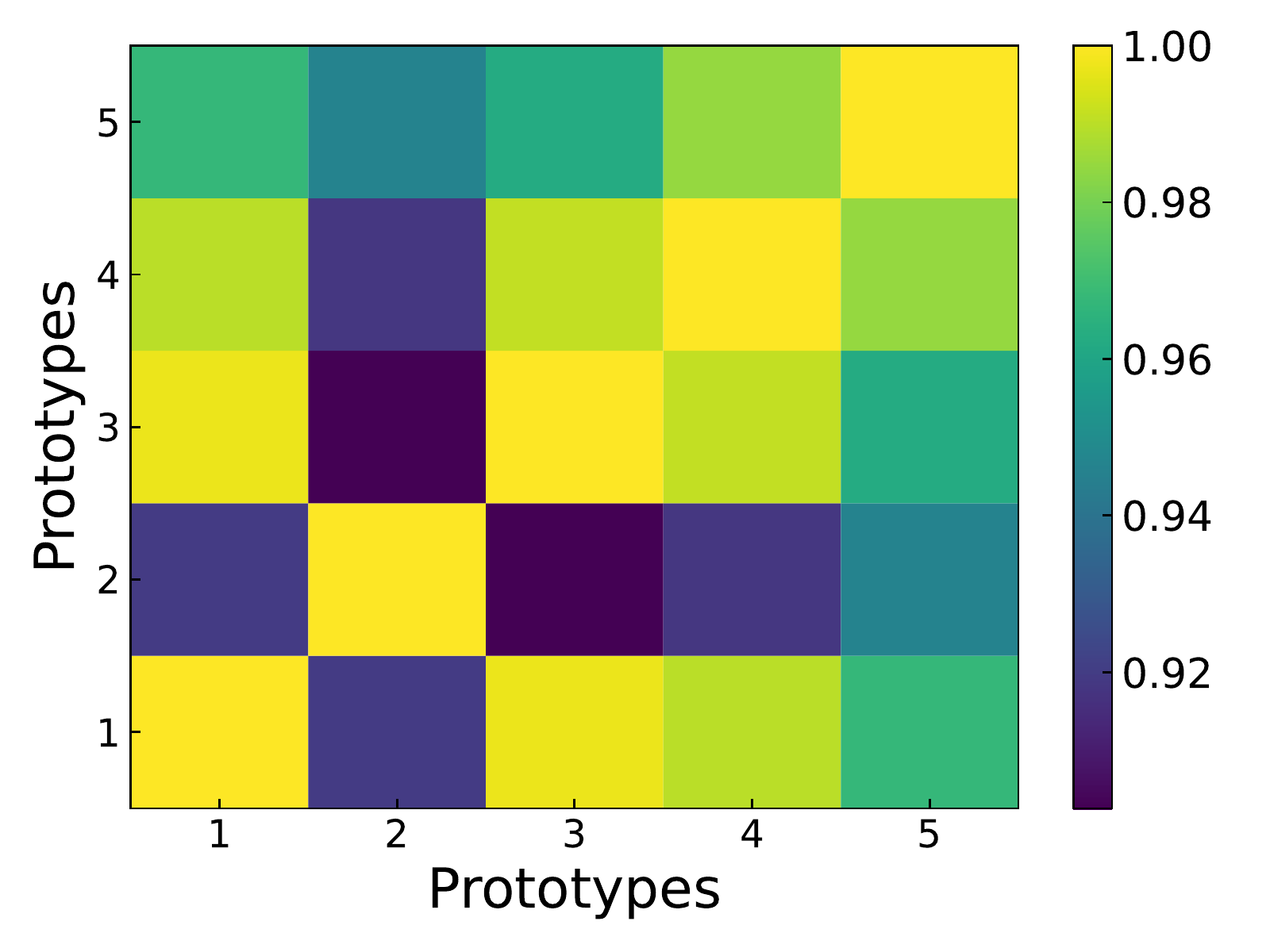}}
    \subfigure[w/o Div Class 1]{\includegraphics[width=0.23\textwidth]{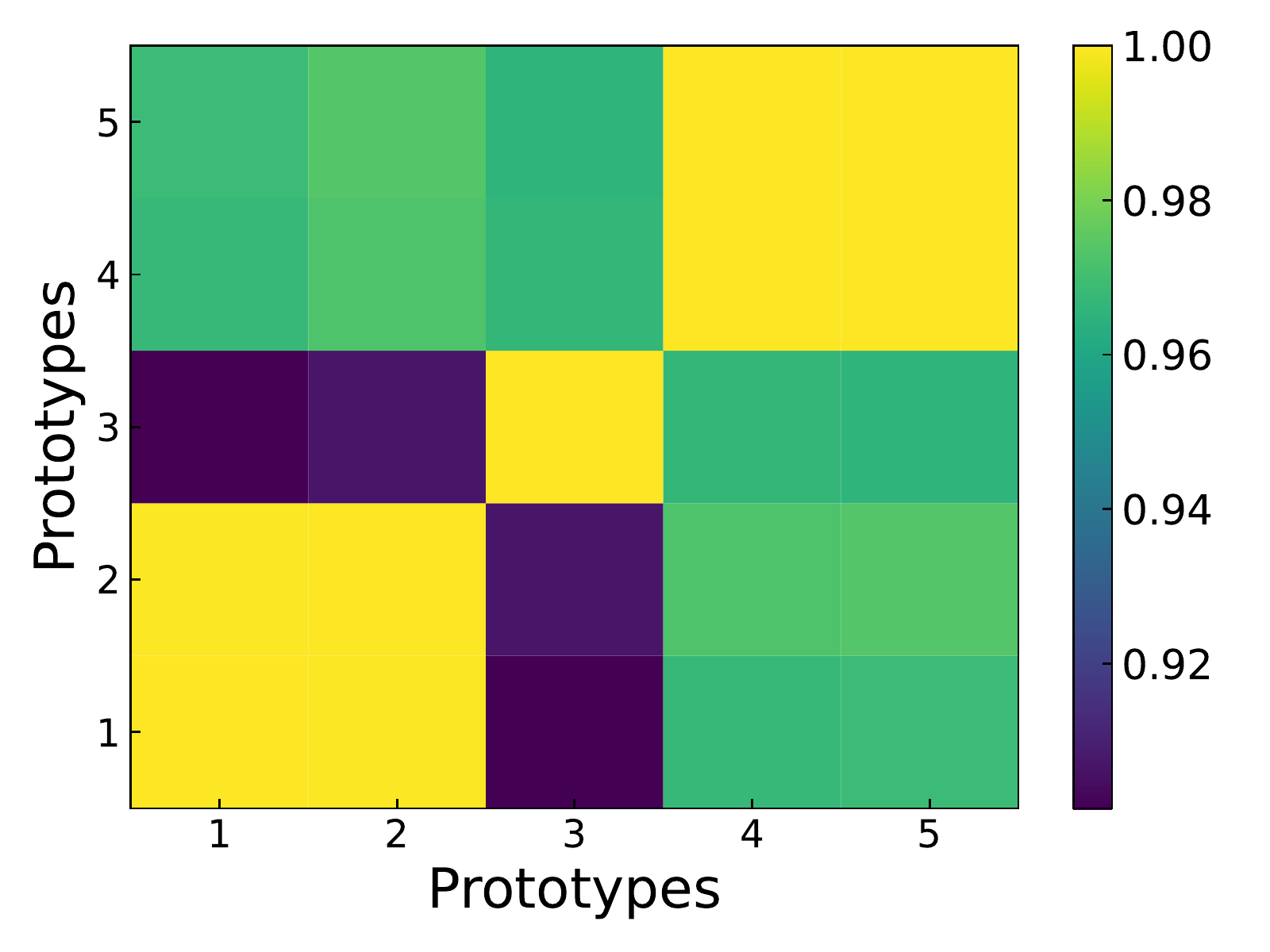}}
    
    \caption{The influence of the diversity regularization in Eq. (\ref{diversity loss}). In the first row, we show the cosine similarities among prototypes learned with the diversity loss. In the Second row, the diversity loss is removed from the learning objective. The cosine similarities among prototypes without diversity regularization are much larger than those with diversity loss. In some extreme cases, the similarities are close to 1, which means the learned prototypes are nearly the same.}
    \label{similarity matrix}
\end{figure}

\begin{figure}[t]
	\centering
	\includegraphics[width=0.4\textwidth]{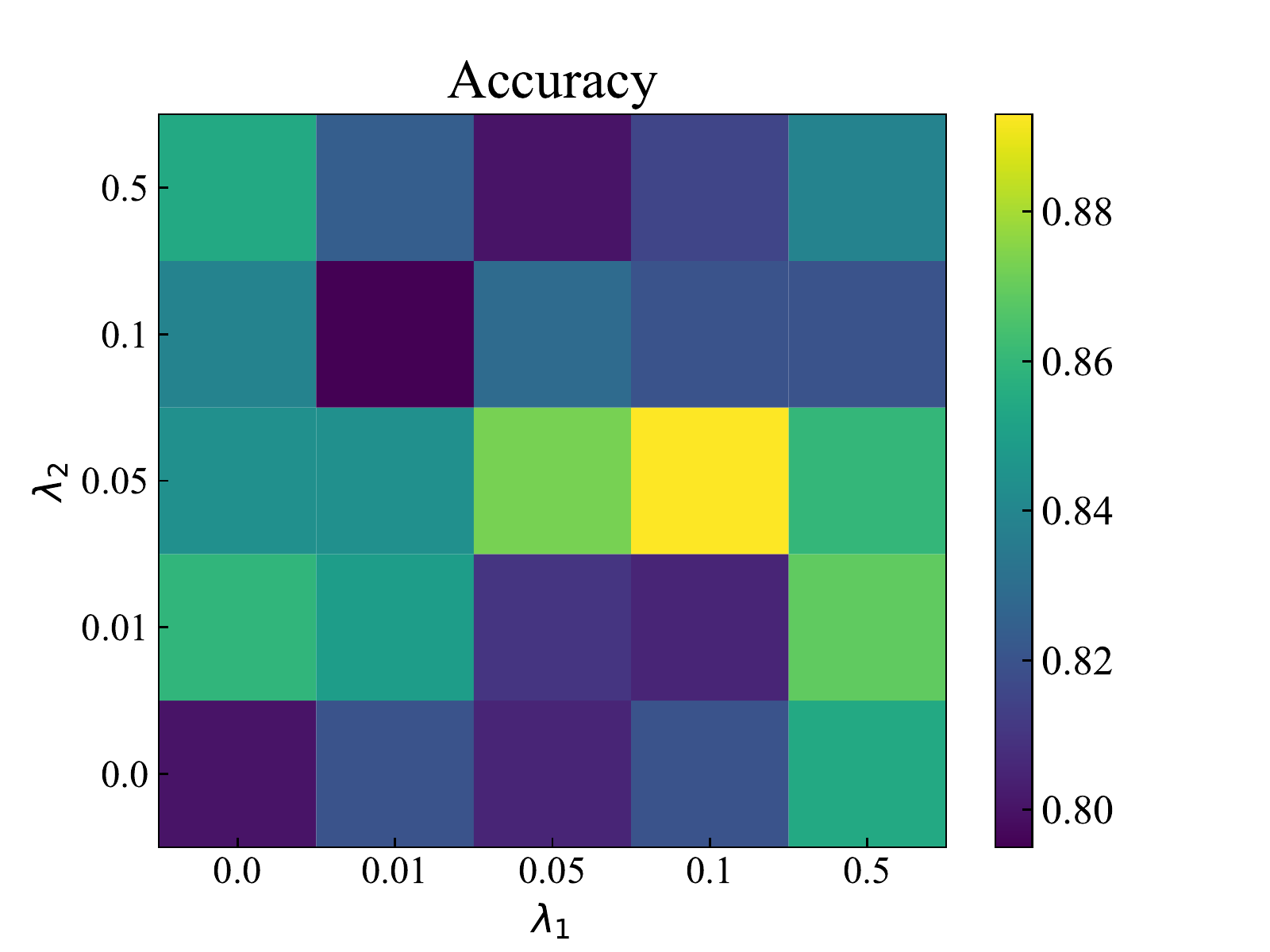}
	\caption{Effects of the cluster and separation losses on BBBP. When $\lambda_1 = 0.10$ and $\lambda_2 = 0.05$, ProtGNN achieves the best performance on BBBP dataset. }
	\label{cluster&separation}
\end{figure}
In Figure \ref{case study 2} and Figure \ref{case study 3}, we show more case studies on BBBP and Graph-Twitter. Note that Graph-Twitter is a 3-class dataset and we show the prototypes for negativeness, neutrality, and positiveness. The input graph in Figure \ref{case study 3} is positive. Our method can effectively capture the key phrase/subgraph leading to positiveness, ``amazing lady gaga I love".
\section{Hyper-parameters Analysis}
In this section, we provide some analysis on hyper-parameters in ProtGNN/ProtGNN+. 
\subsection{Choosing the Number of Prototypes per Class}

We first investigate how would the number of prototypes per class $m$ influence the performance of ProtGNN using BBBP and Graph-Twitter. With the default setting of hyper-parameters, we train ProtGNN with varying $m$. In Figure \ref{m}, we observe that the accuracy of ProtGNN firstly increase dramatically as $m$ increases. Then the increasing slope flattens after $m$ exceeds 5.

Actually, there is one trade-off between accuracy and interpretability when choosing $m$. The accuracy increases when $m$ increases. However, a large number of prototypes makes the model difficult to train and comprehend. In experiments, we choose $m = 5$ since the increasing $m$ only brings marginal improvement to the performance.
\subsection{Influence of Diversity Loss}
We further show the effectiveness of the diversity loss (Eq. \ref{diversity loss}). In Figure \ref{similarity matrix}, we plot the cosine similarity matrices of the learned prototypes on BBBP. The first row are similarity matrices with the diversity loss while the second row without the diversity loss. We can observe that the cosine similarities among prototypes without diversity regularization are much larger than those with diversity loss. In some extreme cases, the similarities are close to 1, which means the learned prototypes are nearly the same. Therefore, the diversity loss can help ProtGNN learn more diverse and evenly distributed prototypes.
\subsection{Influence of the Cluster and Separation Loss}
Here we want to show the influence of $\lambda_1$ and $\lambda_2$ which controls the weights of the cluster loss and separation loss respectively. In Figure \ref{cluster&separation}, we can observe that the cluster and separation constraints play important roles in ProtGNN. When $\lambda_1 = 0.10$ and $\lambda_2 = 0.05$, ProtGNN achieves the best performance on BBBP dataset.

\end{document}